\documentclass[runningheads]{llncs}
\usepackage{graphicx}
\usepackage{amsmath,amssymb} 
\usepackage{color}
\usepackage[width=122mm,left=12mm,paperwidth=146mm,height=193mm,top=12mm,paperheight=217mm]{geometry}
\usepackage{url}      
\usepackage{float}
\usepackage{array}
\usepackage{amsfonts} 
\usepackage{amstext} 
\usepackage{amsmath} 
\usepackage{mathtools}
\usepackage{wrapfig,lipsum,booktabs}
\usepackage{subcaption}
\usepackage[font=small,labelsep=period,belowskip=0pt,aboveskip=4pt]{caption}
\usepackage[dvipsnames]{xcolor}
\DeclarePairedDelimiter\norm{\lVert}{\rVert}%

\DeclareMathOperator*{\argmin}{argmin}

\colorlet{revcol2}{purple!100!black}
\colorlet{revcol}{black}

\newcommand{\Dan}[1]{\textcolor{YellowOrange}{[\textbf{Dan}: #1]}} 

\newcommand{\eg}{e.g.\ }

\newcommand{\parahead}[1]{\setlength{\parindent}{0cm}\par\textbf{#1}:\ }
\newcommand{\tabhead}[1]{\par\textbf{#1}}
\newenvironment{packed_itemize}
{\begin{itemize}
    \setlength{\itemsep}{1pt}
    \setlength{\parskip}{0pt}
    \setlength{\parsep}{0pt}
}{\end{itemize}}

\newcommand{\MU}{\boldsymbol{\mu}}

\newcommand{\filluptopage}[1]{%
  \clearpage
  \loop\ifnum\value{page}<#1\relax
    \null\clearpage
  \repeat
  \loop\ifnum\value{page}=#1\relax
    \null\clearpage
  \repeat
}

\begin{document}
\pagestyle{headings}
\mainmatter
\title{Real-time Joint Tracking of a Hand Manipulating an Object from RGB-D Input}
\titlerunning{Real-time Joint Hand and Object Tracking from RGB-D Input}
\authorrunning{S.\,Sridhar, F.\,Mueller, M.\,Zollh\"ofer, D.\,Casas, A.\,Oulasvirta, C.\,Theobalt}
\author{Srinath Sridhar\textsuperscript{1}~~~~~~Franziska Mueller\textsuperscript{1}
  ~~~~~~Michael Zollh\"ofer\textsuperscript{1}\\Dan Casas\textsuperscript{1}~~~~~
  Antti Oulasvirta\textsuperscript{2}~~~~~Christian Theobalt\textsuperscript{1}
}
%
%
%
\institute{\textsuperscript{1}Max Planck Institute for Informatics~~\textsuperscript{2}Aalto University\\
  \email{\{ssridhar,frmueller,mzollhoef,dcasas,theobalt\}@mpi-inf.mpg.de\\\{antti.oulasvirta\}@aalto.fi}
}
\maketitle
\begin{abstract}
  Real-time simultaneous tracking of hands manipulating and interacting with external objects has many potential applications in augmented reality, tangible computing, and wearable computing.
  However, due to difficult occlusions, fast motions, and uniform hand appearance, jointly tracking hand and object pose is more challenging than tracking either of the two separately.
  Many previous approaches resort to complex multi-camera setups to remedy the occlusion problem and often employ expensive segmentation and optimization steps which makes real-time tracking impossible.
  In this paper, we propose a real-time solution that uses a single commodity RGB-D camera.
  The core of our approach is a 3D articulated Gaussian mixture alignment strategy tailored to hand-object tracking that allows fast pose optimization.
  The alignment energy uses novel regularizers to address occlusions and hand-object contacts.
  For added robustness, we guide the optimization with discriminative part classification of the hand and segmentation of the object.
  We conducted extensive experiments on several existing datasets and introduce a new annotated hand-object dataset.
  Quantitative and qualitative results show the key advantages of our method: speed, accuracy, and robustness.
\end{abstract}

\section{Introduction}
\begin{figure}[t]
  \centering
  \includegraphics[width=1.0\linewidth]{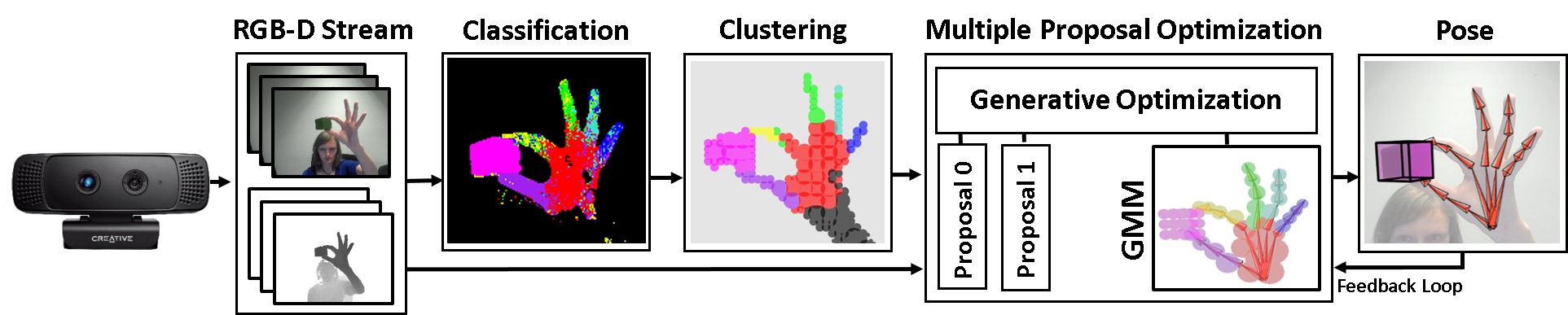}
  \caption{
  We perform classification of the input into object and hand parts. The hand and object are tracked using 3D articulated Gaussian mixture alignment
}
  \label{fig:pipeline}
\end{figure}

The human hand exhibits incredible capacity for manipulating external objects via gripping, grasping, touching, pointing, caging, and throwing.
We can use our hands with apparent ease, even for subtle and complex motions, and with remarkable speed and accuracy.
However, this dexterity also makes it hard to track a hand in close interaction with objects.
While a lot of research has explored real-time tracking of hands or objects in isolation, real-time hand-object tracking remains unsolved.
It is inherently more challenging due to the higher dimensionality of the problem, additional occlusions, and difficulty in disambiguating hand from object.
A fast, accurate, and robust solution based on a minimal camera setup is a precondition for many new and important applications in vision-based input to computers,
including virtual and augmented reality, teleoperation, tangible computing, and wearable computing.
In this paper, we present a \textbf{real-time} method to \textbf{simultaneously track} a hand and the manipulated object.
We support tracking objects of \textbf{different shapes, sizes}, and \textbf{colors}.
\begin{figure}
  \centering
  \includegraphics[width=\textwidth]{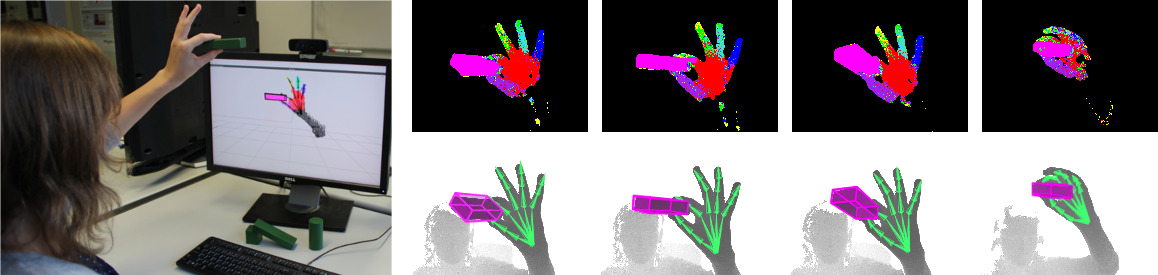}
  \caption{Proposed real-time hand-object tracking approach: we use a single commodity depth camera (\textit{left}) to classify (\textit{top}) and track the articulation of a hand and the rigid body motion of a manipulated object (\textit{bottom})}
  \label{fig:teaser}
\end{figure}
Previous work has employed setups with multiple cameras~\cite{ballan_motion_2012,oikonomidis_full_2011} to limit the influence of occlusions
which restricts use to highly controlled setups.
Many methods that exploit dense depth and color measurements from commodity RGB-D cameras~\cite{hamer_tracking_2009,kyriazis_physically_2013,kyriazis_scalable_2014} have been proposed.
However, these methods use expensive segmentation and optimization steps that make interactive performance hard to attain.
At the other end of the spectrum, discriminative one-shot methods (for tracking only hands) often suffer from temporal instability~\cite{KeskinKKA11,TangCTK14,XuChe:iccv13}.
Such approaches have also been applied to estimate hand pose under object occlusion~\cite{romero_hands_2010}, but the object is not tracked simultaneously.
In contrast, the approach proposed here is the first to track hand and object motion simultaneously at real-time rates using only a single commodity RGB-D camera (see Fig.~\ref{fig:teaser}).
Building on recent work in single hand tracking and 3D pointset registration, we propose a 3D articulated Gaussian mixture alignment strategy tailored to hand-object tracking.
Gaussian mixture alignment aligns two Gaussian mixtures and has been successfully used in 3D pointset registration~\cite{jian2011robust}.
It can be interpreted as a generalization of ICP and does not require explicit, error-prone, and computationally expensive correspondence search~\cite{campbell2016gogma}.
Previous methods have used articulated 2.5D Gaussian mixture alignment formulations~\cite{FastHandTracker_CVPR2015} that are discontinuous.
This leads to tracking instabilities because 3D spatial proximity is not considered.
We also introduce additional novel regularizers that consider occlusions and enforce contact points between fingers and objects analytically.
Our combined energy has a closed form gradient and allows for fast and accurate tracking.
For an overview of our approach see Figure~\ref{fig:pipeline}.
To further increase robustness and allow for recovery of the generative tracker, we guide the optimization using a multi-layer random forest hand part classifier.
We use a variational optimization strategy that optimizes two different hand-object tracking energies simultaneously (multiple proposals) and then selects the better solution.
The main contributions are:
\begin{packed_itemize}
\item A 3D articulated Gaussian mixture alignment approach for jointly tracking hand and object accurately.
\item Novel contact point and occlusion objective terms that were motivated by the physics of grasps, and can handle difficult hand-object interactions.
\item A multi-layered classification architecture to segment hand and object, and classify hand parts in RGB-D sequences.
\item An extensive evaluation on public datasets as well as a new, fully annotated dataset consisting of diverse hand-object interactions.
\end{packed_itemize}

\section{Related Work}
\paragraph{\textbf{Single Hand Tracking}}
Single hand tracking has received a lot of attention in recent years with discriminative and generative methods being the two main classes of methods.
Discriminative methods for monocular RGB tracking index into a large database of poses or learn a mapping from image to pose space~\cite{AthitsosS03,Wu2000}.
However, accuracy and temporal stability of these methods are limited.
Monocular generative methods optimize pose of more sophisticated 3D or 2.5D hand models by optimizing an alignment energy~\cite{Heap1996,Bray2004,deLaGorce2011}.
Occlusions and appearance ambiguities are less problematic with multi-camera setups~\cite{ballan_motion_2012}. 
\cite{Wang:2013} use a physics-based approach to optimize the pose of a hand using silhouette and color constraints at slow non-interactive frame rates.
\cite{sridhar2013} use multiple RGB cameras 
and a single depth camera to track single hand poses in near real-time by combining generative tracking and finger tip detection. 
More lightweight setups with a single depth camera are preferred for many interactive applications.  
Among single camera methods, examples of discriminative methods are based on decision forests for hand part labeling \cite{KeskinKKA11}, on a latent regression forest in combination with a coarse-to-fine search \cite{TangCTK14}, fast hierarchical pose regression~\cite{Sun_2015_CVPR}, or Hough voting~\cite{XuChe:iccv13}. Real-time performance is feasible, but temporal instability remains an issue. 
\cite{OikonomidisBMVC11} generatively track a hand by optimizing a depth and appearance-based alignment metric with particle swarm optimization (PSO).
A real-time generative tracking method with a physics-based solver was
proposed in~\cite{Melax:2013}. The stabilizaton of real-time articulated ICP based on a learned subspace prior on hand poses was used in~\cite{htrack_sgp15}.
Template-based non-rigid deformation tracking of arbitrary objects in real-time 
from RGB-D was shown in~\cite{zollhoefer2014deformable}, very simple unoccluded hand poses can be tracked. 
Combining generative and discriminative tracking enables recovery from some tracking failures~\cite{handpose_chi2015,tzionas_capturing_2014,sridhar2013}.
\cite{FastHandTracker_CVPR2015} showed real-time single hand tracking from depth using generative pose optimization under detection constraints.  
Similarly, reinitialization of generative estimates via finger tip detection~\cite{Qian_2014_CVPR}, multi-layer discriminative reinitialization~\cite{handpose_chi2015}, or joints detected with convolutional networks is feasible~\cite{Tompson:2014}.  
\cite{Tang_ICCV_2015} employ hierarchical sampling from partial pose distributions and a final hypothesis selection based on a generative energy. 
None of the above methods is able to track interacting hands and objects simultaneously and in non-trivial poses in real-time. 

\paragraph{\textbf{Tracking Hands in Interaction}}
Tracking two interacting hands, or a hand and a manipulated object, is a much harder problem.
The straightforward combination of methods for object tracking, e.g.\ \cite{badami_depth-enhanced,tejani_latent-class_2014}, and hand tracking does not lead to satisfactory
solutions, as only a combined formulation can methodically exploit mutual constraints between object and hand.   
\cite{Wang:2011} track two well-separated hands from stereo by efficient pose retrieval and 
IK refinement.
In~\cite{Oikon:2012} two hands in interaction are tracked at 4\,Hz with an RGB-D camera by 
optimizing a generative depth and image alignment measure. 
Tracking of interacting hands from multi-view video at slow non-interactive runtimes was shown in~\cite{ballan_motion_2012}.
They use generative pose optimization supported by salient point detection. 
The method in~\cite{htrack_sgp15} can track very simple two hand interactions with little occlusion. 
Commercial solutions, e.g. Leap Motion \cite{leapmotion} and NimbleVR\cite{nimblevr}, fail if two hands interact closely or interact with an object.
In \cite{oikonomidis_full_2011}, a marker-less method based on a generative pose optimization of a combined hand-object model is proposed. They explicitly model collisions, but need multiple RGB cameras.
In~\cite{hamer_tracking_2009} the most likely pose is found through belief propagation using part-based trackers. 
This method is robust under occlusions, but does not explicitly track the object.
A temporally coherent nearest neighbor search tracks the hand manipulating an object in~\cite{romero_hands_2010}, but not the object, in real time. Results are prone to temporal jitter.
\cite{kyriazis_physically_2013} perform frame-to-frame tracking of hand and objects from RGB-D using physics-based optimization.
This approach has a slow non-interactive runtime.
An ensemble of Collaborative Trackers (ECT) for RGB-D based multi-object and multiple hand tracking is used in~\cite{kyriazis_scalable_2014}. Their accuracy is high, but runtime is far from real-time.
\cite{Pham_2015_CVPR} infer contact forces from a tracked hand interacting with an object at slow non-interactive runtimes.
\cite{BMVC2015_123} and \cite{Tzionas_ICCV_2015} propose methods for in-hand RGB-D object scanning.
Both methods use known generative methods to track finger contact points to support ICP-like shape scanning.
Recently, \cite{tzionas2015capturing} introduced a method for tracking hand-only, hand-hand, and hand-object (we include a comparison with this method).
None of the above methods can track the hand and the manipulated object in \emph{real-time} in non-trivial poses from a \emph{single depth camera} view, which is what our approach achieves.

\paragraph{\textbf{Model-based Tracking Approaches}}
A common representation for model tracking are meshes~\cite{ballan_motion_2012,htrack_sgp15}.
Other approaches use primitives~\cite{kyriazis_scalable_2014,Qian_2014_CVPR},
quadrics~\cite{stenger2001model}, 2.5D Gaussians~\cite{FastHandTracker_CVPR2015}, or Gaussian mixtures~\cite{jian2011robust}.
Gaussian mixture alignment has been successfully used in rigid pointset registration~\cite{jian2011robust}.
In contrast, we propose a 3D \emph{articulated} Gaussian mixture alignment strategy.
\cite{YeM:CVPR2014} relate template and data via a probabilistic formulation and use EM to compute the best fit.
Different from our approach, they only model the template as a Gaussian mixture.

\begin{figure}[t]
	\centering
	\includegraphics[width=0.85\linewidth]{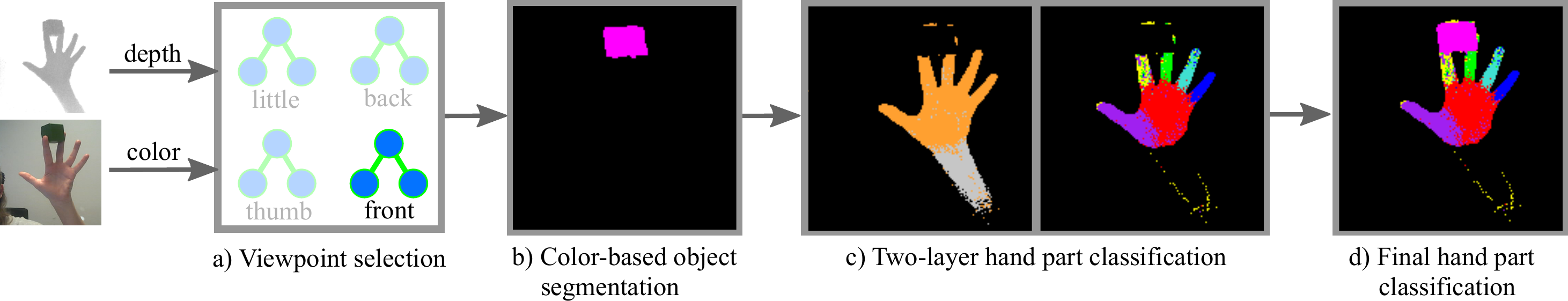}
	\caption{Three stage hand part classification: Stage 1: Viewpoint selection, stage 2: color-based object segmentation, and stage 3: two-layer hand part classification}
	\label{fig:strategies}
\end{figure}

\section{Discriminative Hand Part Classification} \label{sec:segmentation}
As a preprocessing step, we classify depth pixels as hand or object, and further into hand parts.
The obtained labeling is later used to guide the generative pose optimization.
Our part classification strategy is based on a two-layer random forest that takes occlusions into account.
Classification is based on a three step pipeline (see Fig.~\ref{fig:strategies}).
Input is the color $\mathcal{C}_t$ and depth $\mathcal{D}_t$ frames captured by the RGB-D sensor.
We first perform hand-object segmentation based on color cues to remove the object from the depth map.
Afterwards, we select a suitable two-layer random forest to obtain the classification.
The final output per pixel is a part probability histogram that encodes the class likelihoods.
Note, object pixel histograms are set to an object class probability of 1.
The forests are trained based on a set of training images that consists of real hand motions re-targeted to a virtual hand model to generate synthetic data from multiple viewpoints.
A virtual object is automatically inserted in the scene to simulate occlusions.
To this end, we randomly sample uniform object positions between the thumb and one other finger and prune implausible poses based on intersection tests.

\paragraph{\textbf{Viewpoint Selection}}
We trained two-layer forests for hand part classification from different viewpoints.
Four cases are distinguished: observing the hand from the front, back, thumb and little finger sides.
We select the forest that best matches the hand orientation computed in the last frame.
The selected two-layer forest is then used for hand part classification.
\paragraph{\textbf{Color-Based Object Segmentation}}
As a first step, we segment out the object from the captured depth map $\mathcal{D}_t$.
Similar to many previous hand-object tracking approaches~\cite{OikonomidisBMVC11}, we use the color image $\mathcal{C}_t$ in combination with an HSV color segmentation strategy.
As we show in the results, we are able to support objects with different colors.
Object pixels are removed to obtain a new depth map $\mathcal{\hat D}_t$, which we then feed to the next processing stage.

\paragraph{\textbf{Two-Layer Hand Part Classification}}
We use a two-layer random forest for hand part classification.
The first layer classifies hand and arm pixels while the second layer uses the hand pixels and further classifies them into one of several distinct hand parts.
Both layers are per-pixel classification forests~\cite{Shotton:2011}.
The hand-arm classification forest is trained on $N = 100k$ images with diverse hand-object poses.
For each of the four viewpoints a random forest is trained on $N=38k$ images.
The random forests are based on three trees, each trained on a random distinct subset.
In each image, 2000 example foreground pixels are chosen.
Split decisions at nodes are based on 100 random feature offsets and 40 thresholds.
Candidate features are a uniform mix of unary and binary depth difference features~\cite{Shotton:2011}.
Nodes are split as long as the information gain is sufficient and the maximum tree depth of 19 (21 for hand-arm forest) has not been reached.
On the first layer, we use 3 part labels: 1 for hand, 1 for arm and 1 to represent the background.
On the second layer, classification is based on 7 part labels: 6 for the hand parts, and 1 for the background. 
We use one label for each finger and one for the palm, see Fig.~\ref{fig:strategies}c.
We use a cross-validation procedure to find the best hyperparameters.
On the disjoint test set, the hand-arm forest has a classification accuracy of 65.2\%.
The forests for the four camera views had accuracies of 59.8\% (front), 64.7\% (back), 60.9\% (little), and 53.5\% (thumb).

\section{Gaussian Mixture Model Representation}
Joint hand-object tracking requires a representation that allows for accurate tracking, is robust to outliers, and enables fast pose optimization.
Gaussian mixture alignment, initially proposed for rigid pointset alignment (e.g.\ \cite{jian2011robust}), satisfies all these requirements.
It features the advantages of ICP-like methods, without requiring a costly, error-prone correspondence search.
We extend this approach to 3D articulated Gaussian mixture alignment tailored to hand-object tracking.
Compared to our 3D formulation, 2.5D \cite{FastHandTracker_CVPR2015} approaches are discontinuous.
This causes instabilities, since the spatial proximity between model and data is not fully considered.
We quantitatively show this for hand-only tracking (Section \ref{sec:results}).

\section{Unified Density Representation}\label{sec:model}

We parameterize the articulated motion of the human hand using a kinematic skeleton with $|\mathcal{X}_h| = 26$ degrees of freedom (DOF).
Non-rigid hand motion is expressed based on $20$ joint angles in twist representation.
The remaining $6$ DOFs specify the global rigid transform of the hand with respect to the root joint.
The manipulated object is assumed to be rigid and its motion is parameterized using $|\mathcal{X}_o| = 6$ DOFs.
In the following, we deal with the hand and object in a unified way.
To this end, we refer to the vector of all unknowns as $\mathcal{X}$.
For pose optimization, both the input depth as well as the scene (hand and object) are expressed as $3$D Gaussian Mixture Models (GMMs).
This allows for fast and analytical pose optimization.
We first define the following generic probability density distribution $\mathcal{M}(\mathbf{x}) = \sum_{i=1}^{K}{w_i \mathcal{G}_i(\mathbf{x}|\MU_i,\sigma_i)}$ at each point $\mathbf{x} \in \mathbb{R}^3$ in space.
This mixture contains $K$ unnormalized, isotropic Gaussian functions $\mathcal{G}_i$ with mean $\MU_i \in \mathbb{R}^3$ and variance $\sigma_i^2 \in \mathbb{R}$.
In the case of the model distribution, the positions of the Gaussians are parameterized by the unknowns $\mathcal{X}$.
For the hand, this means each Gaussian is being rigidly rigged to one bone of the hand.
The probability density is defined and non-vanishing over the whole domain $\mathbb{R}^3$.

\paragraph{\textbf{Hand and Object Model}}
The three-dimensional shape of the hand and object is represented in a similar fashion as probability density distributions $\mathcal{M}_{h}$ and $\mathcal{M}_{o}$, respectively.
We manually attach $N_h= 30$ Gaussian functions to the kinematic chain of the hand to model its volumetric extent.
Standard deviations are set such that they roughly correspond to the distance to the actual surface.
The object is represented by automatically fitting a predefined number $N_o$ of Gaussians to its spatial extent, such that the one standard deviation spheres model the objects volumetric extent.
$N_o$ is a user defined parameter which can be used to control the trade-off between tracking accuracy and runtime performance.
We found that $N_o \in [12, 64]$ provides a good trade-off between speed and accuracy for the objects used in our experiments.
We refer to the combined hand-object distribution as $\mathcal{M}_{s}$, with $N_s = N_h + N_o$ Gaussians.
Each Gaussian is assigned to a class label $l_i$ based on its semantic location in the scene.
Note, the input GMM is only a model of the visible surface of the hand/object.
Therefore, we incorporate a visibility factor $f_i \in [0, 1]$ ($0$ completely occluded, $1$ completely visible) per Gaussian.
This factor is approximated by rendering an occlusion map with each Gaussian as a circle (radius equal to its standard deviation).
The GMM is restricted to the visible surface by setting $w_i=f_i$ in the mixture.
These operations are performed based on the solution of the previous frame $\mathcal{X}_{old}$.

\paragraph{\textbf{Input Depth Data}}
We first perform bottom-up hierarchical quadtree clustering of adjacent pixels with similar depth to convert the input to the density based representation.
We cluster at most $(2^{(4-1)})^2=64$ pixels, which corresponds to a maximum tree depth of $4$.
Clustering is performed as long as the depth variance in the corresponding subdomain is smaller than $\epsilon_{cluster}=30$ mm.
Each leaf node is represented as a Gaussian function $\mathcal{G}_i$ with $\boldsymbol \mu_i$ corresponding to the $3$D center of gravity of the quad and $\sigma_i^2=(\frac{a}{2})^2$, where $a$ is the backprojected side length of the quad.
Note, the mean $\boldsymbol \mu_i \in \mathbb{R}^3$ is obtained by backprojecting the $2$D center of gravity of the quad based on the computed average depth and displacing by $a$ in camera viewing direction to obtain a representation that matches the model of the scene.
In addition, each $\mathcal{G}_i$ stores the probability $p_i$ and index $l_i$ of the best associated semantic label.
We obtain the best label and its probability by summing over all corresponding per-pixel histograms obtained in the classification stage.
Based on this data, we define the input depth distribution $\mathcal{M}_{d_h}(\mathbf{x})$ for the hand and $\mathcal{M}_{d_o}(\mathbf{x})$ for the object.
The combined input distribution $\mathcal{M}_{d}(\mathbf{x})$ has $N_d = N_{d_o} + N_{d_h}$ Gaussians.
We set uniform weights $w_i=1$ based on the assumption of equal contribution.
$N_d$ is much smaller than the number of pixels leading to real-time hand-object tracking.

\section{Multiple Proposal Optimization} \label{sec:optimization}
We optimize for the best pose $\mathcal{X}^*$ using two proposals $\mathcal{X}_i^*,~ i \in \{0, 1\}$ that are computed by minimizing two distinct hand-object tracking energies:
\begin{equation}
\mathcal{X}_0^* = \argmin_{\mathcal{X}}E_{align}(\mathcal{X}),~\mathcal{X}_1^* = \argmin_{\mathcal{X}}E_{label}(\mathcal{X})~.
\end{equation}
$E_{align}$ leverages the depth observations and the second energy $E_{label}$ incorporates the discriminative hand part classification results.
In contrast to the optimization of the sum of the two objectives, this avoids failure due to bad classification and ensures fast recovery.
For optimization, we use analytical gradient descent ($10$ iterations per proposal, adaptive step length)~\cite{StollHGST11}.
We initialize based on the solution of the previous frame $\mathcal{X}_{old}$.
Finally, $\mathcal{X}^*$ is selected as given below, where we slightly favor ($\lambda = 1.003$) the label proposal to facilitate fast pose recovery:
\begin{equation}
  \mathcal{X}^*  = 
  \begin{cases}
    \mathcal{X}_1^* & \text{if } \big( E_{val}(\mathcal{X}_1^*) < \lambda  E_{val}(\mathcal{X}_0^*) \big) \\
    \mathcal{X}_0^* & \text{otherwise}\\
  \end{cases}~.
\end{equation}
The energy $E_{val}(\mathcal{X}) = E_a(\mathcal{X}) + w_p E_p(\mathcal{X})$ is designed to select the proposal that best explains the input, while being anatomically correct.
Therefore, it considers spatial alignment to the input depth map $E_a$ and models anatomical joint angle limits $E_p$, see Section \ref{sec:tracking}.
In the following, we describe the used energies in detail.

\section{Hand-Object Tracking Objectives} \label{sec:tracking}
Given the input depth distribution $\mathcal{M}_d$, we want to find the 3D model $\mathcal{M}_s$ that best explains the observations by varying the corresponding parameters $\mathcal{X}$.
We take inspiration from methods with slow non-interactive runtimes that used related 3D implicit shape models for full-body pose tracking~\cite{plankers_articulated_2003,Kurman2013}, but propose a new efficient tracking objective tailored for real-time hand-object tracking.
In contrast to previous methods, our objective operates in $3$D (generalization of ICP), features an improved way of incorporating the discriminative classification results, and incorporates two novel regularization terms.
Together, this provides for a better, yet compact, representation that allows for fast analytic pose optimization on the CPU.
To this end, we define the following two objective functions.
The first energy $E_{align}$ measures the alignment with the input:
\begin{equation}
  E_{align}(\mathcal{X}) = E_{a} + w_p E_{p} + w_t E_{t} + w_c E_c + w_o E_o~.
\end{equation}
The second energy $E_{label}$ incorporates the classification results:
\begin{align}
  E_{label}(\mathcal{X}) = E_{a} + w_s E_{s}+ w_p E_{p}~.
\end{align}
The energy terms consider spatial alignment $E_{a}$, semantic alignment $E_{s}$, anatomical plausibility $E_{p}$, temporal smoothness $E_{t}$, contact points $E_c$, and object-hand occlusions $E_o$, respectively.
The priors in the energies are chosen such that they do not hinder the respective alignment objectives.
All parameters $w_p=0.1$, $w_t=0.1$, $w_s=3 \cdot 10^{-7}$, $w_c=5 \cdot 10^{-7}$ and $w_o=1.0$ have been empirically determined and stay fixed for all experiments.
We optimize both energies simultaneously using a multiple proposal based optimization strategy and employ a winner-takes-all strategy (see Section \ref{sec:optimization}).
We found empirically that using two energy functions resulted in better pose estimation and recovery from failures than using a single energy with all terms.
In the following, we give more details on the individual components.

\paragraph{\textbf{Spatial Alignment}}
We measure the alignment of the input density function $\mathcal{M}_d$ and our scene model $\mathcal{M}_s$ based on the following alignment energy:
\begin{equation}
  \small
  \medmuskip=-1mu
  \thinmuskip=-1mu
  \thickmuskip=-1mu
  E_a(\mathcal{X}) =  \int_\Omega \Big[ \big(\mathcal{M}_{d_{h}}(\mathbf{x}) - \mathcal{M}_h(\mathbf{x}) \big)^2\\ + 
  \big(\mathcal{M}_{d_{o}}(\mathbf{x}) - \mathcal{M}_o(\mathbf{x}) \big)^2\Big]d\mathbf{x}~.
\end{equation}
It measures the alignment between the two input and two model density distributions at every point in space $\mathbf{x} \in \Omega$.
Note, this $3$D formulation leads to higher accuracy results (see Section \ref{sec:results}) than a $2.5$D \cite{FastHandTracker_CVPR2015} formulation.
 
\paragraph{\textbf{Semantic Alignment}}
In addition to the alignment of the distributions, we also incorporate semantic information in the label energy $E_{label}$.
In contrast to~\cite{FastHandTracker_CVPR2015}, we incorporate uncertainty based on the best class probability.
We use the following least-squares objective to enforce semantic alignment:
\begin{equation}
E_s(\mathcal{X}) = \sum_{i=1}^{N_s} \sum_{j=1}^{N_d} \alpha_{i, j}\cdot ||\boldsymbol \mu_i - \boldsymbol \mu_j||^2_2~.
\end{equation}
Here, $\boldsymbol \mu_i$ and $\boldsymbol \mu_j$  are the mean of the $i^{th}$ model and  the $j^{th}$ image Gaussian, respectively.
The weights $\alpha_{i, j}$ switch attraction forces between similar parts on and between different parts off:
\begin{equation}
  \alpha_{i, j}  =
  \begin{cases}
    0 & \text{if (} l_i \neq l_j ) \text{ or } (d_{i,j} > r_{max}) \\
    (1 - \frac{d_{i,j}}{r_{max}}) \cdot p_i& \text{else}\\
  \end{cases}.
\end{equation}
Here, $d_{i,j} = || \boldsymbol \mu_i - \boldsymbol \mu_j||_2$ is the distance between the means.
$l_i$ is the part label of the most likely class, $p_i$ its probability and $r_{max}$ a cutoff value.
We set $r_{max}$ to $30$cm.
$l_i$ can be one of $8$ labels: $6$ for the hand parts, $1$ for object and $1$ for background.
We consider all model Gaussians, independent of their occlusion weight, to facilitate fast pose recovery of previously occluded regions.

\paragraph{\textbf{Anatomical Plausibility}}
The articulated motion of the hand is subject to anatomical constraints.
We account for this by enforcing soft-constraints on the joint angles $\mathcal{X}_h$ of the hand:
\begin{equation}
  E_{p}(\mathcal{X})  = \sum_{x_i \in \mathcal{X}_h}
  \begin{cases}
    0 & \text{if } x_i^{l} \le x_i \le x_i^u \\
    \norm{x_i  -  x_i^l}^2 & \text{if }  x_i  < x_i^l \\
    \norm{x_i^u  - x_i}^2 & \text{if }   x_i > x_i^u\\
  \end{cases}.
\end{equation}
Here, $\mathcal{X}_h$ are the DOFs corresponding to the hand, and $x_i^l$ and $x_i^u$ are the lower and upper joint limit that corresponds to the $i^{th}$ DOF of the kinematic chain.

\paragraph{\textbf{Temporal Smoothness}}
We further improve the smoothness of our tracking results by incorporating a temporal prior into the energy.
To this end, we include a soft constraint on parameter change to enforce constant speed:
\begin{equation}
E_{t}(\mathcal{X}) = \norm{ \nabla \mathcal{X} -  \nabla \mathcal{X}^{(t-1)}}_2^2~.
\end{equation}
Here, $\nabla \mathcal{X}^{(t-1)}$ is the gradient of parameter change at the previous time step.

\paragraph{\textbf{Contact Points}}
We propose a novel contact point objective, specific to the hand-object tracking scenario:
\begin{equation}
E_{c}(\mathcal{X}) = \sum_{(k, l, t_d) \in \mathcal{T}}{}~ \Big( ||\boldsymbol \mu_k - \boldsymbol \mu_l||^2 - t_{d}^2 \Big)^2~.
\end{equation}
Here, $(k, l, t_d) \in \mathcal{T}$ is a detected touch constraint.
It encodes that the fingertip Gaussian with index $k$ should have a distance of $t_d$ to the object Gaussian with index $l$.
We detect the set of all touch constraints $\mathcal{T}$ based on the last pose $\mathcal{X}_{old}$.
A new touch constraint is added if a fingertip Gaussian is closer to an object Gaussian than the sum of their standard deviations.
We then set $t_d$ to this sum.
This couples hand pose and object tracking leading to more stable results.
A contact point is active until the distance between the two Gaussians exceeds the release threshold $\delta_R$.
Usually $\delta_R > t_d$ to avoid flickering.

\paragraph{\textbf{Occlusion Handling}}

No measurements are available in occluded hand regions.
We stabilize the hand movement in such regions using a novel occlusion prior:
\begin{equation}
E_{o}(\mathcal{X}) = \sum_{i=0}^{N_h} \sum_{j \in \mathcal{H}_i}~ (1-\hat{f}_i) \cdot ||x_j - x_j^{old}||^2_2~.
\end{equation}
Here, $\mathcal{H}_i$ is the set of all DOFs that are influenced by the $i$-th Gaussian.
The global rotation and translation is not included.
The occlusion weights $\hat{f}_i \in [0,1]$ are computed similar to $f_i$ ($0$ occluded, $1$ visible).
This prior is based on the assumption that occuded regions move consistently with the rest of the hand.
\section{Experiments and Results}\label{sec:results}
\begin{figure}
  \begin{subfigure}[b]{\textwidth}
    \centering
    \includegraphics[width=0.19\linewidth]{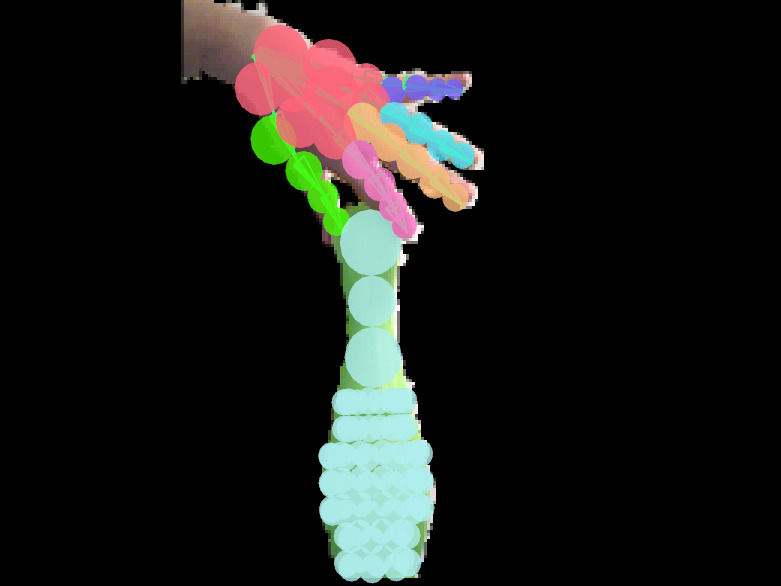}
    \includegraphics[width=0.19\linewidth]{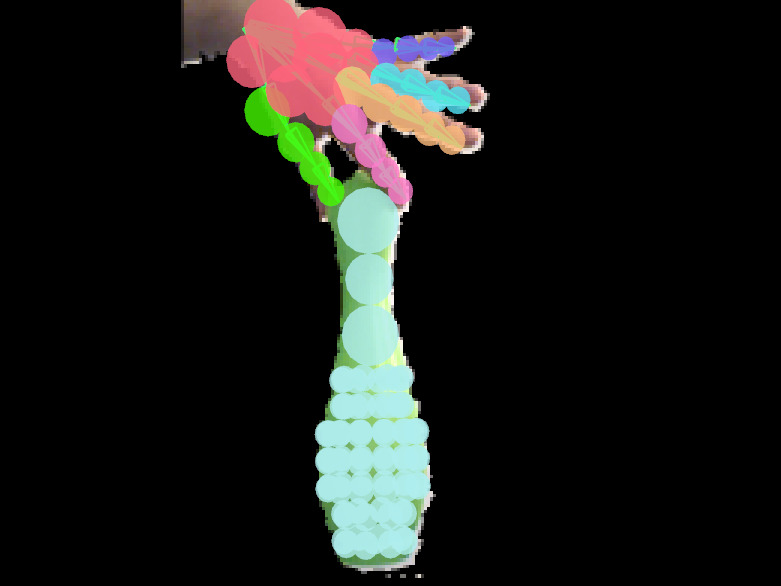}
    \includegraphics[width=0.19\linewidth]{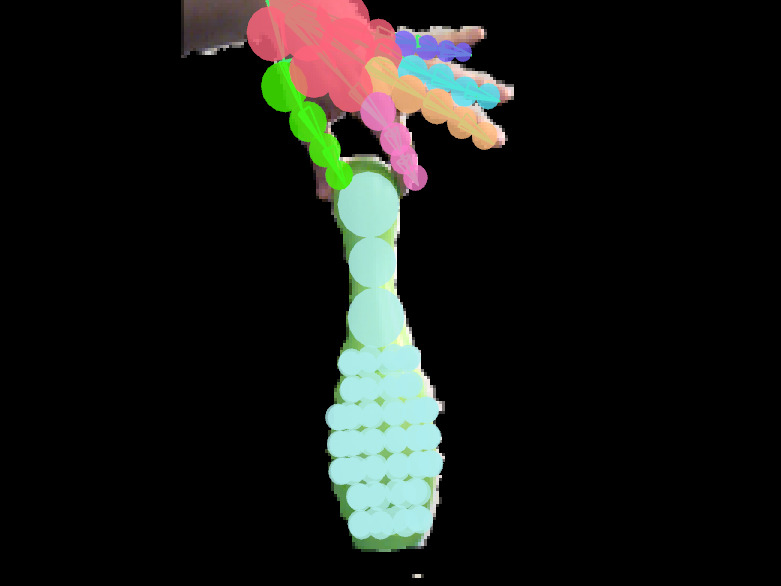}
    \includegraphics[width=0.19\linewidth]{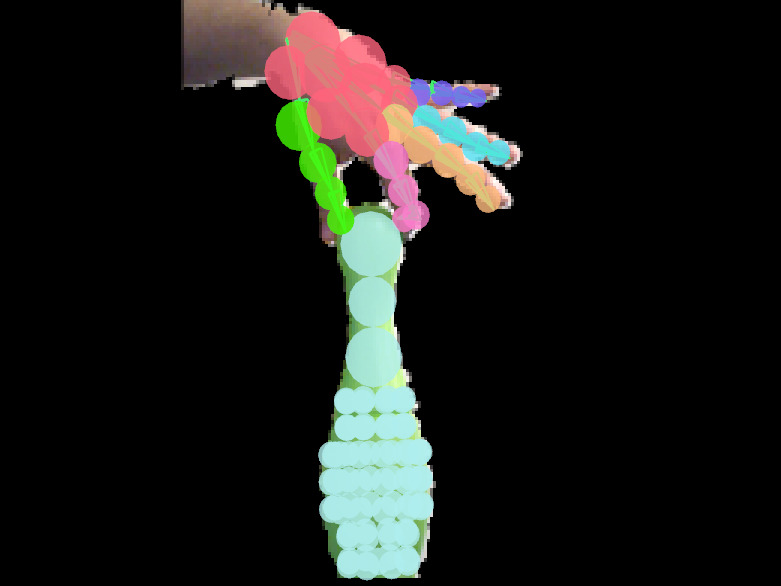}
    \includegraphics[width=0.19\linewidth]{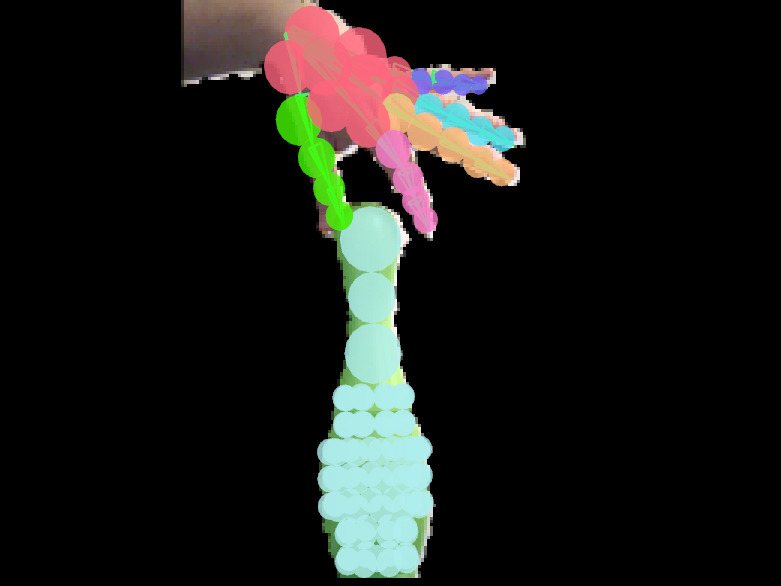}
  \end{subfigure}		
  \caption{Subset of tracked frames on the dataset of \cite{Tzionas_ICCV_2015}. Our method can handle
    objects with \textbf{varying sizes, colors, and different hand dimensions}. Here we show how even a complex shape like a bowling pin can be approximated using only a few tens of Gaussians
    }
  \label{fig:sota_qual}
\end{figure}

We evaluate and compare our method on more than \textbf{15 sequences} spanning 3 public datasets, which have been recorded with 3 different 
RBG-D cameras.
Additional live sequences (see Fig.~\ref{fig:sota_qual} and supplementary materials) show that our method handles fast object and finger motion, difficult occlusions and fares well even if two hands are present in the scene.
Our method supports commodity RGB-D sensors like the \emph{Creative Senz3D}, \emph{Intel RealSense F200}, and \emph{Primesense Carmine}.
We rescale depth and color to resolutions of 320$\times$240 and 640$\times$480 respectively, and capture at 30\,Hz.
Furthermore, we introduce a new hand-object tracking benchmark dataset with ground truth fingertip and object annotations.

\paragraph{\textbf{Comparison to the State-of-the-Art}}
We quantitatively and qualitatively evaluate on two publicly available hand-object
datasets \cite{tzionas2015capturing,Tzionas_ICCV_2015} (see Fig.~\ref{fig:sota_qual} and also supplementary material).
Only one dataset (IJCV~\cite{tzionas2015capturing}) contains ground truth joint annotations.
We test on 5 rigid object sequences from IJCV.
We track the right hand only, but our method works even when multiple hands are present.
Ground truth annotations are provided for 2D joint positions, but not object pose.
Our method achieves a fingertip pixel error of \textbf{8.6px}, which is comparable (difference of only 2px) to that reported for the slower method of \cite{tzionas2015capturing}.
This small difference is well within the uncertainty of manual annotation and sensor noise.
Note, our approach runs over 60 times faster, while producing visual results that are on par (see Fig.~\ref{fig:sota_qual}).
We also track the dataset of ~\cite{Tzionas_ICCV_2015} (see also Fig.~\ref{fig:sota_qual}).
While they solve a different problem (offline in-hand scanning), it shows that our real-time method copes well with different shaped objects (\eg bowling pin, bottle, etc.) under occlusion.

\begin{figure}[t]
  \centering
  \begin{subfigure}[t]{0.48\textwidth}
    \includegraphics[height=2.5cm]{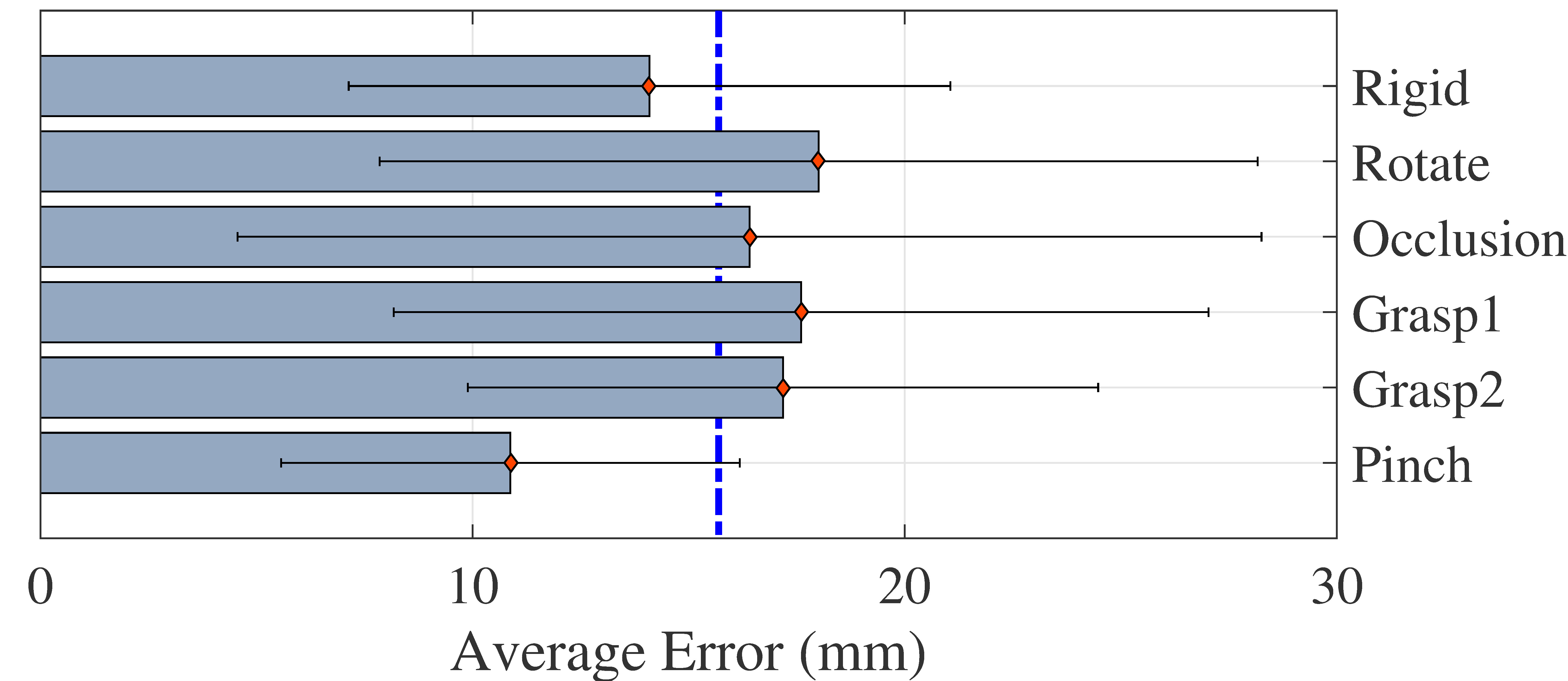}
    \caption{We achieve low errors on each of the 6 sequences in our new benchmark dataset}
    \label{fig:accuracy}
  \end{subfigure}
  ~
  \begin{subfigure}[t]{0.42\textwidth}
    \includegraphics[height=2.5cm]{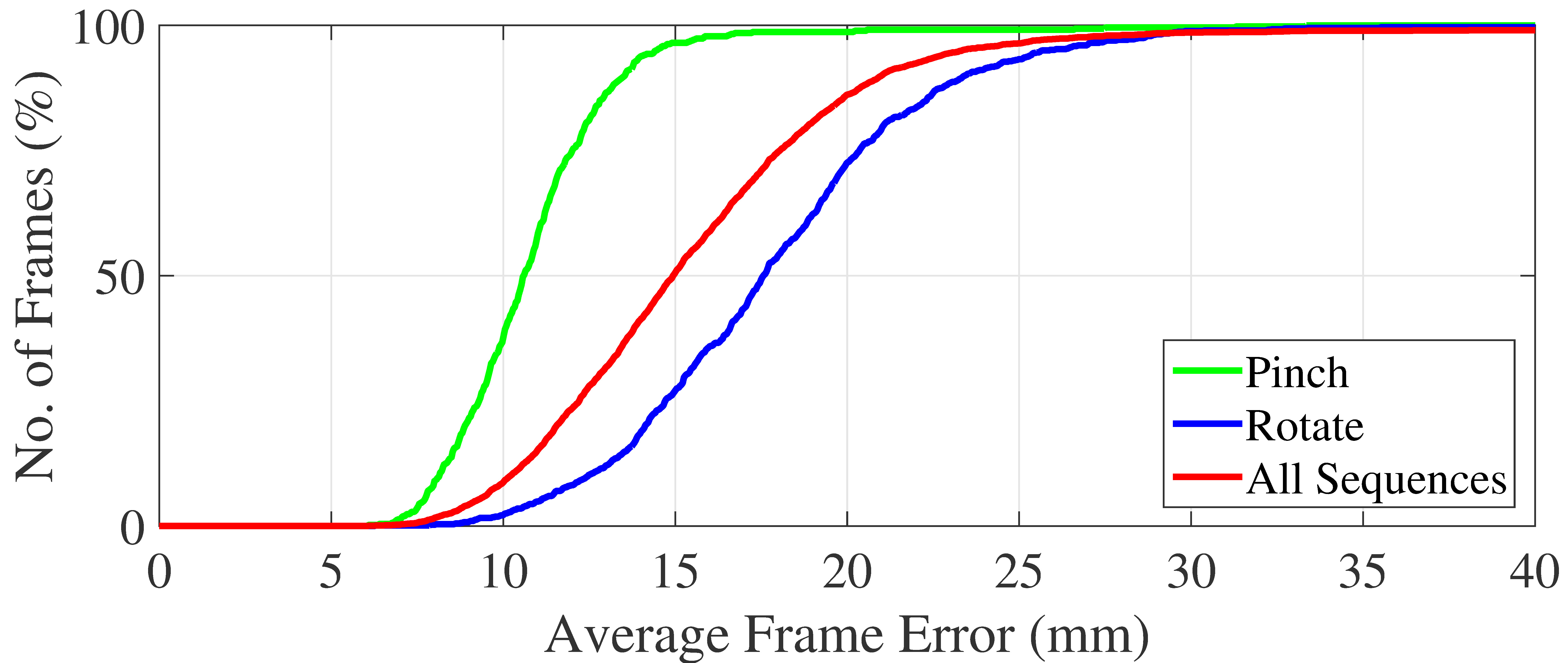}
    \caption{Tracking consistency of the best, worst and average case} 
    \label{fig:consistency}
  \end{subfigure}
  \caption{Quantitative hand-object tracking evaluation on ground truth data. The object contributes a higher error}
\end{figure}

\paragraph{\textbf{New Benchmark Dataset}}
With the aforementioned datasets, evaluation of object pose is impossible due to missing object annotations.
We therefore introduce, to our knowledge, the first dataset\footnote{\url{http://handtracker.mpi-inf.mpg.de/projects/RealtimeHO/}} that contains ground truth for \textbf{both} fingertip positions and object pose.
It contains 6 sequences of a hand manipulating a cuboid (2 different sizes) in different hand-object configurations and grasps.
We manually annotated pixels on the depth image to mark 5 fingertip positions, and 3 cuboid corners.
In total, we provide 3014 frames with ground truth annotations.
%
%
As is common in the literature~\cite{handpose_chi2015,TangCTK14,FastHandTracker_CVPR2015,Qian_2014_CVPR,htrack_sgp15}, we use the average
3D Euclidean distance $E$ between estimated and ground truth positions as the error measure (see supplementary document for details).
Occluded fingertips are excluded on a per-frame basis from the error computation.
If one of the annotated corners of the cuboid is occluded, we exclude it from that frame.
In Fig.~\ref{fig:accuracy} we plot the average error over all frames of the 6 sequences.
Our method has an average error (for both hand and object) of \textbf{15.7\,mm}.
Over all sequences, the average error is always lower than 20\,mm with standard deviations under 12\,mm.
Average error is an indicator of overall performance, but does not indicate how consistent the tracker performs.
Fig.~\ref{fig:consistency} shows that our method tracks almost all frames with less than 30\,mm error.
\emph{Rotate} has the highest error, while \emph{Pinch} performs best with almost all frames below 20\,mm.
Table~\ref{tab:ablative} shows the errors for hand and object separately.
Both are in the same order of magnitude.
\begin{table}[t]
  \centering
  \caption{Average error (mm) for hand and object tracking in our dataset}
  \begin{tabular}{|c|c|c|c|c|c|c||c|}
    \hline
    \tabhead{} & \emph{Rigid} & \emph{Rotate} & \emph{Occlusion} & \emph{Grasp1} & \emph{Grasp2} & \emph{Pinch} & \tabhead{Overall (mm)}\\
    \hline
    Fingertips   & 14.2 & 16.3 & 17.5 & 18.1 & 17.5 & 10.3  & \textbf{15.6}\\
    \hline
    Object        & 13.5 & 26.8 & 11.9 & 15.3 & 15.7 & 13.9  & \textbf{16.2}\\
    \hline
    Combined ($E$) & 14.1 & 18.0 & 16.4 & 17.6 & 17.2 & 10.9  & \textbf{15.7}\\
    \hline
  \end{tabular} 
  \label{tab:ablative}
\end{table}
\begin{figure}[t]
  \centering
  \raisebox{3mm}{\begin{subfigure}[b]{0.03\textwidth}
      \includegraphics[trim=10cm 0.2cm -0.5cm 5cm, clip=true, width=\linewidth]{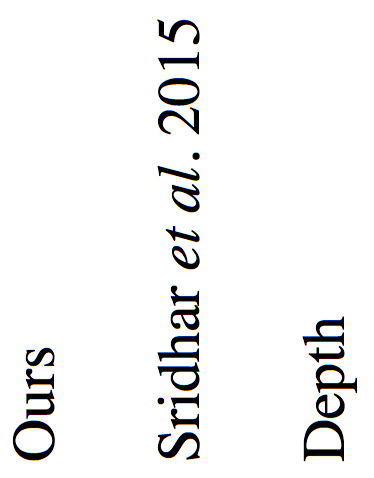}
  \end{subfigure}}
  \begin{subfigure}[b]{0.13\textwidth}
    \includegraphics[trim=1cm 0.5cm 2.5cm 2cm, clip=true,width=\linewidth]{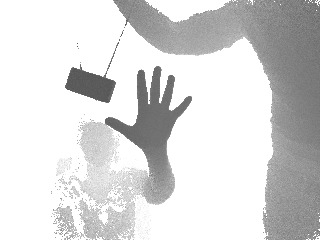}
  \end{subfigure}
  \begin{subfigure}[b]{0.13\textwidth}
    \includegraphics[trim=1cm 0.5cm 2.5cm 2cm, clip=true,width=\linewidth]{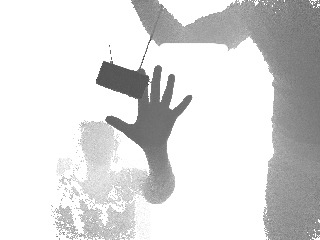}
  \end{subfigure}
  \begin{subfigure}[b]{0.13\textwidth}
    \includegraphics[trim=1cm 0.5cm 2.5cm 2cm, clip=true,width=\linewidth]{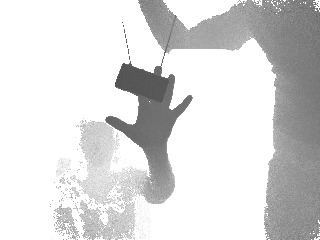}
  \end{subfigure}
  \begin{subfigure}[b]{0.13\textwidth}
    \includegraphics[trim=1cm 0.5cm 2.5cm 2cm, clip=true,width=\linewidth]{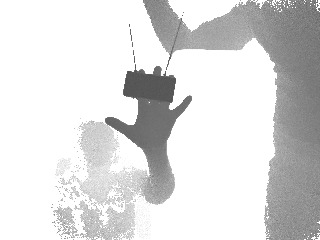}
  \end{subfigure}
  \begin{subfigure}[b]{0.13\textwidth}
    \includegraphics[trim=1cm 0.5cm 2.5cm 2cm, clip=true,width=\linewidth]{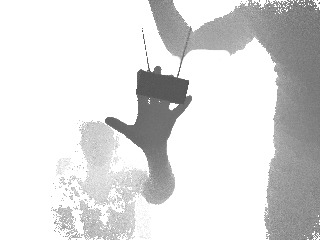}
  \end{subfigure}
  \begin{subfigure}[b]{0.13\textwidth}
    \includegraphics[trim=1cm 0.5cm 2.5cm 2cm, clip=true,width=\linewidth]{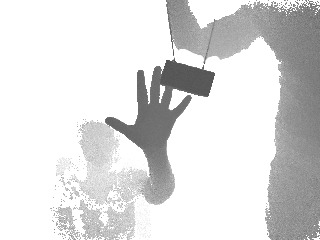}
  \end{subfigure}
  \begin{subfigure}[b]{0.13\textwidth}
    \includegraphics[trim=1cm 0.5cm 2.5cm 2cm, clip=true,width=\linewidth]{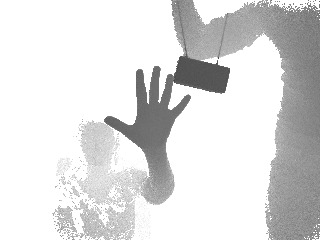}
  \end{subfigure}	
  \\[0.1cm]
  \raisebox{7mm}{\begin{subfigure}[b]{0.03\textwidth}
      \includegraphics[trim=-0.5cm 0cm 10cm 8.5cm, clip=true, width=\linewidth]{content/images/comparison_cvpr15/labels.jpg}
    \end{subfigure}}
  \begin{subfigure}[b]{0.13\textwidth}
    \includegraphics[width=\linewidth]{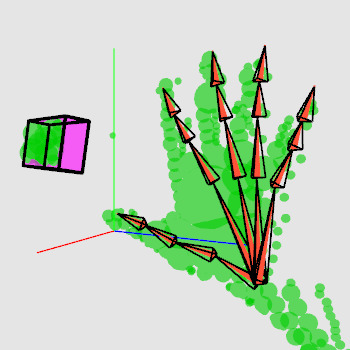}
  \end{subfigure}
  \begin{subfigure}[b]{0.13\textwidth}
    \includegraphics[width=\linewidth]{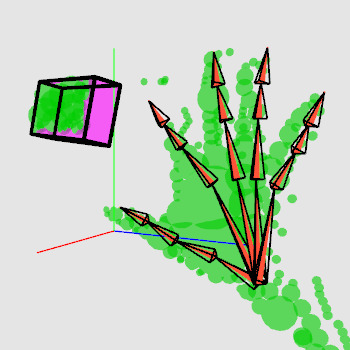}
  \end{subfigure}
  \begin{subfigure}[b]{0.13\textwidth}
    \includegraphics[width=\linewidth]{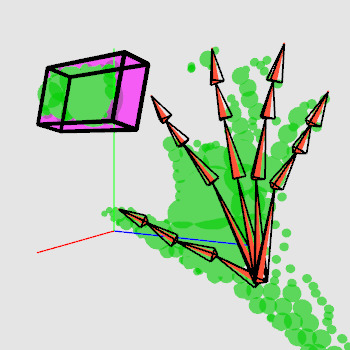}
  \end{subfigure}
  \begin{subfigure}[b]{0.13\textwidth}
    \includegraphics[width=\linewidth]{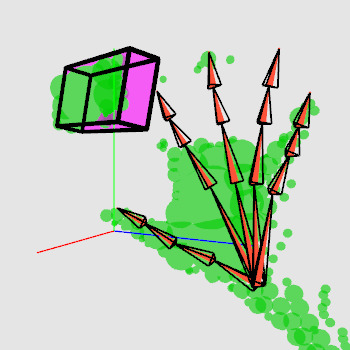}
  \end{subfigure}
  \begin{subfigure}[b]{0.13\textwidth}
    \includegraphics[width=\linewidth]{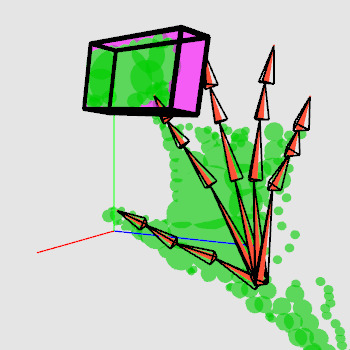}
  \end{subfigure}	
  \begin{subfigure}[b]{0.13\textwidth}
    \includegraphics[width=\linewidth]{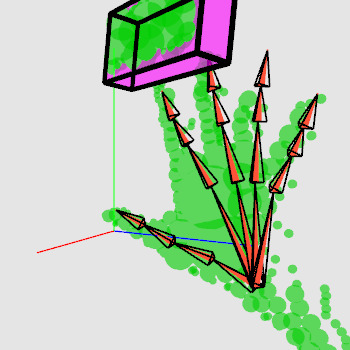}
  \end{subfigure}			
  \begin{subfigure}[b]{0.13\textwidth}
    \includegraphics[width=\linewidth]{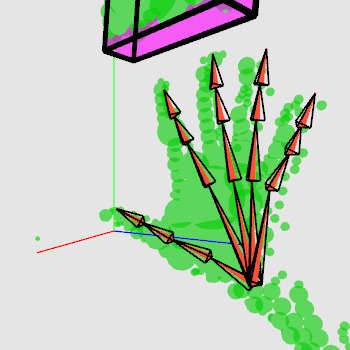}
  \end{subfigure}	
  \\[0.1cm]
  \raisebox{1mm}{\begin{subfigure}[b]{0.03\textwidth}
      \includegraphics[trim=4.5cm 0cm 4.5cm 0cm, clip=true, width=\linewidth]{content/images/comparison_cvpr15/labels.jpg}
    \end{subfigure}}
  \begin{subfigure}[b]{0.13\textwidth}
    \includegraphics[width=\linewidth]{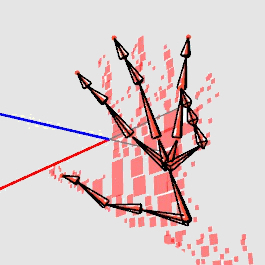}	
  \end{subfigure}
  \begin{subfigure}[b]{0.13\textwidth}
    \includegraphics[width=\linewidth]{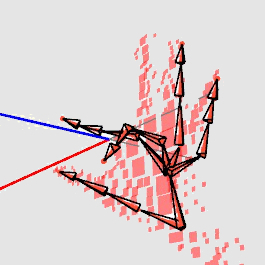}	
  \end{subfigure}	
  \begin{subfigure}[b]{0.13\textwidth}
    \includegraphics[width=\linewidth]{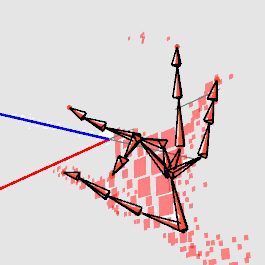}
  \end{subfigure}
  \begin{subfigure}[b]{0.13\textwidth}
    \includegraphics[width=\linewidth]{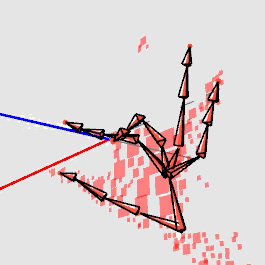}	
  \end{subfigure}	
  \begin{subfigure}[b]{0.13\textwidth}
    \includegraphics[width=\linewidth]{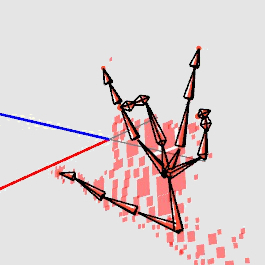}
  \end{subfigure}	
  \begin{subfigure}[b]{0.13\textwidth}
    \includegraphics[width=\linewidth]{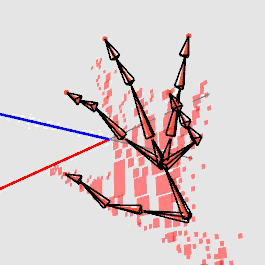}	
  \end{subfigure}	
  \begin{subfigure}[b]{0.13\textwidth}
    \includegraphics[width=\linewidth]{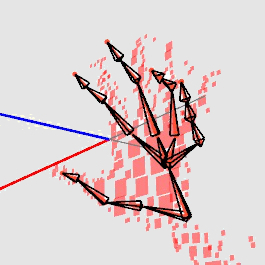}
  \end{subfigure}
  \caption{\textit{Top row:} Input depth, an object occludes the hand. \textit{Middle row:} Result of our approach (different viewpoint).
  Our approach succesfully tracks the hand under heavy occlusion.
  \textit{Bottom row:} Result of~\cite{FastHandTracker_CVPR2015} shows catastrophic failure (object pixels were removed for fairness)
}
  \label{fig:occlusion}
\end{figure}

\paragraph{\textbf{Ablative Analysis}}
Firstly, we show that the articulated 3D Gaussian mixture alignment formulation is superior (even for tracking only hand) to the 2.5D formulation of \cite{FastHandTracker_CVPR2015}.
On the Dexter dataset~\cite{sridhar2013}, \cite{FastHandTracker_CVPR2015} report an average fingertip error of \textbf{19.6\,mm}.
In contrast, our method (\textbf{without} any hand-object specific terms) is consistently better with an average of \textbf{17.2\,mm} (maximum improvement is \textbf{5\,mm} on 2 sequences).
This is a result of the continuous articulated 3D Gaussian mixture alignment energy, a generalization of ICP, which considers 3D spatial proximity between Gaussians.

\setlength\intextsep{0pt}
\begin{wrapfigure}{lh!}{0.45\textwidth}
\centering
\includegraphics[width=0.45\textwidth]{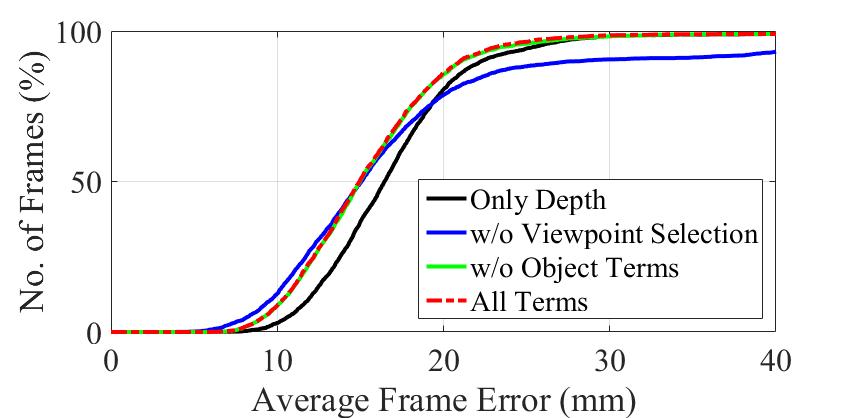}
\caption{\label{fig:ablative} Ablative analysis}
\end{wrapfigure}
Secondly, we show that the average error on our hand-object dataset is worse without viewpoint selection, semantic alignment, occlusion handling, and contact points term.
Fig.~\ref{fig:ablative} shows a consistency plot with different components of the energy disabled.
Using only the data term often results in large errors.
The errors are even larger without viewpoint selection.
The semantic alignment, occlusion handling, and contact points help improve \protect{\textbf{robustness}} of tracking results and \textbf{recovery} from failures.
Fig.~\ref{fig:occlusion} shows that \cite{FastHandTracker_CVPR2015} clearly fails when fingers are occluded.
Our hand-object specific terms are more robust to these difficult occlusion cases while achieving real-time performance.

\paragraph{\textbf{Runtime Performance}}
All experiments were performed on an Intel Xeon E5-1620 CPU with $16$\,GB memory and an NVIDIA GTX~980~Ti.
The stages of our approach take on average: 4\,ms for preprocessing, 4\,ms for part classification, 2\,ms for depth clustering, and 20-30 ms for pose optimization using two proposals.
We achieve real-time performance of 25-30\,Hz.
Multi-layer random forests ran on the GPU while all other algorithm parts ran multithreaded on a CPU.
\begin{figure}
  \centering
  \begin{subfigure}[b]{0.48\textwidth}
  	\centering
    \includegraphics[trim=3.5cm 2.2cm 2.5cm 2cm, clip=true, width=0.31\linewidth]{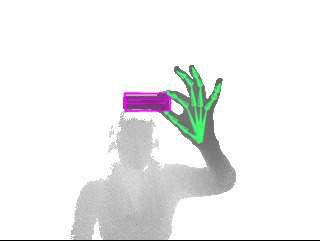}
    \includegraphics[trim=3.5cm 2.2cm 2.5cm 2cm, clip=true, width=0.31\linewidth]{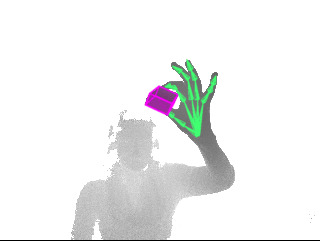}
    \includegraphics[trim=3.5cm 2.2cm 2.5cm 2cm, clip=true, width=0.31\linewidth]{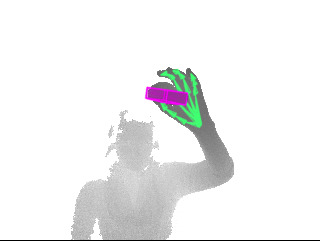}
  \end{subfigure}
  \hspace{0.2cm}		
  \begin{subfigure}[b]{0.48\textwidth}
  	\centering
    \includegraphics[trim=2.5cm 1cm 3.5cm 3cm, clip=true, width=0.31\linewidth]{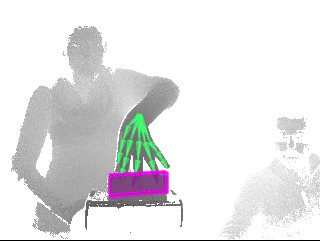}
    \includegraphics[trim=2.5cm 1cm 3.5cm 3cm, clip=true, width=0.31\linewidth]{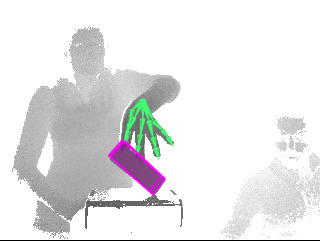}
    \includegraphics[trim=2.5cm 1cm 3.5cm 3cm, clip=true, width=0.31\linewidth]{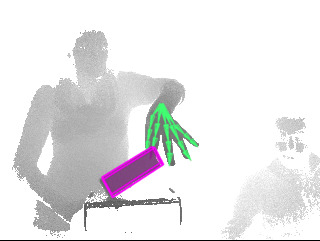}
  \end{subfigure}		
  \\
  \begin{subfigure}[b]{0.48\textwidth}
  	\centering
    \includegraphics[trim=3.5cm 1.5cm 1cm 1cm, clip=true, width=0.31\linewidth]{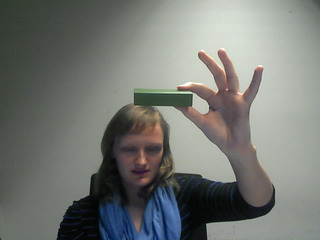}
    \includegraphics[trim=3.5cm 1.5cm 1cm 1cm, clip=true, width=0.31\linewidth]{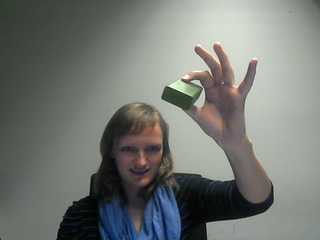}
    \includegraphics[trim=3.5cm 1.5cm 1cm 1cm, clip=true, width=0.31\linewidth]{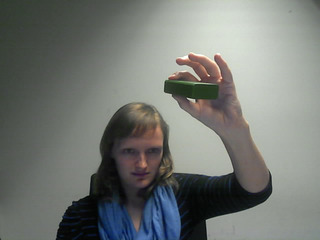}
    \caption{\textit{Rotate} sequence from our dataset}
  \end{subfigure}
  \hspace{0.2cm}	
  \begin{subfigure}[b]{0.48\textwidth}
  	\centering
    \includegraphics[trim=2.5cm 0.5cm 3.2cm 3cm, clip=true, width=0.31\linewidth]{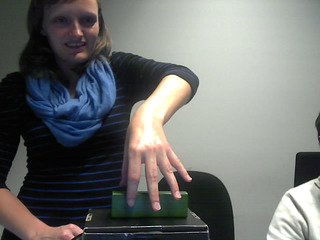}
    \includegraphics[trim=2.5cm 0.5cm 3.2cm 3cm, clip=true, width=0.31\linewidth]{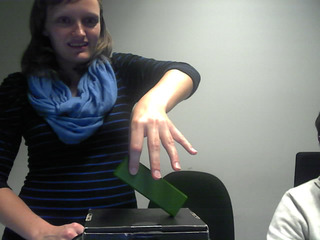}
    \includegraphics[trim=2.5cm 0.5cm 3.2cm 3cm, clip=true, width=0.31\linewidth]{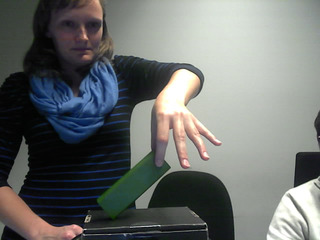}
    \caption{\textit{Grasp2} sequence from our dataset}
  \end{subfigure}
 \\
  \begin{subfigure}[b]{\textwidth}
  	\centering
    \includegraphics[trim=1.5cm 2.8cm 4.2cm 1cm, clip=true, width=0.16\linewidth]{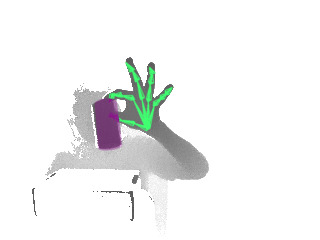}
    \includegraphics[trim=1.5cm 2.8cm 4.2cm 1cm, clip=true, width=0.16\linewidth]{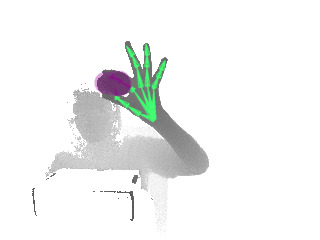}
    \includegraphics[trim=1.5cm 2.8cm 4.2cm 1cm, clip=true, width=0.16\linewidth]{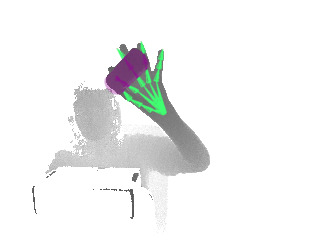}
    \includegraphics[trim=3cm 2.5cm 3cm 1.8cm, clip=true, width=0.16\linewidth]{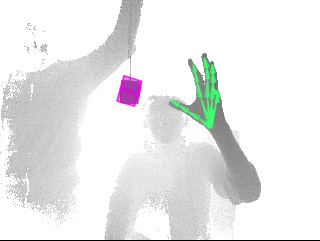}
    \includegraphics[trim=3cm 2.5cm 3cm 1.8cm, clip=true, width=0.16\linewidth]{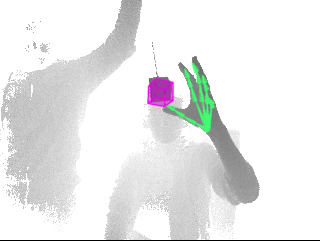}
    \includegraphics[trim=3cm 2.5cm 3cm 1.8cm, clip=true, width=0.16\linewidth]{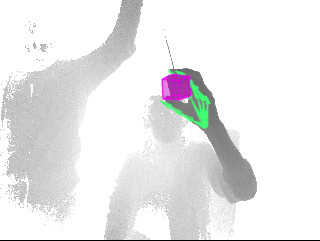}
 \\
    \includegraphics[width=0.16\linewidth]{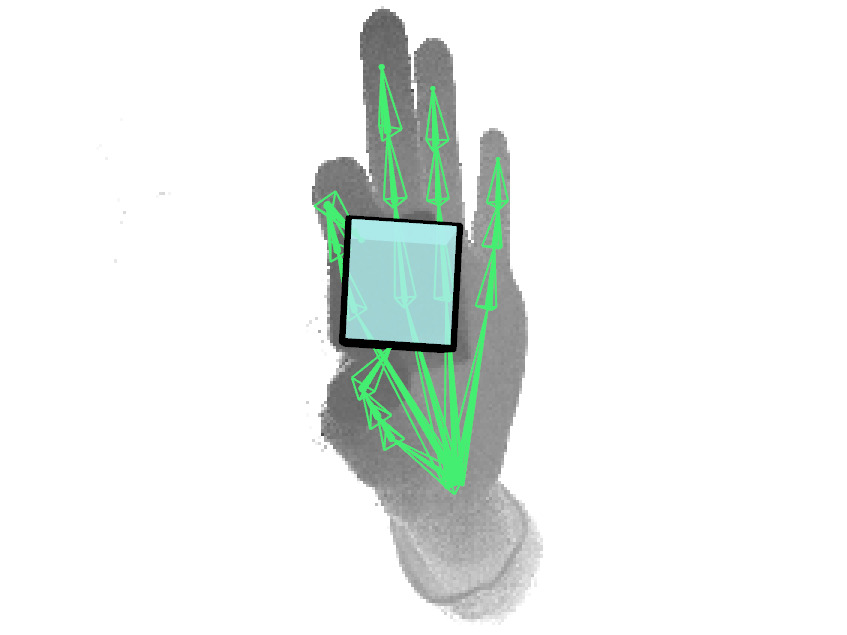}
    \includegraphics[width=0.16\linewidth]{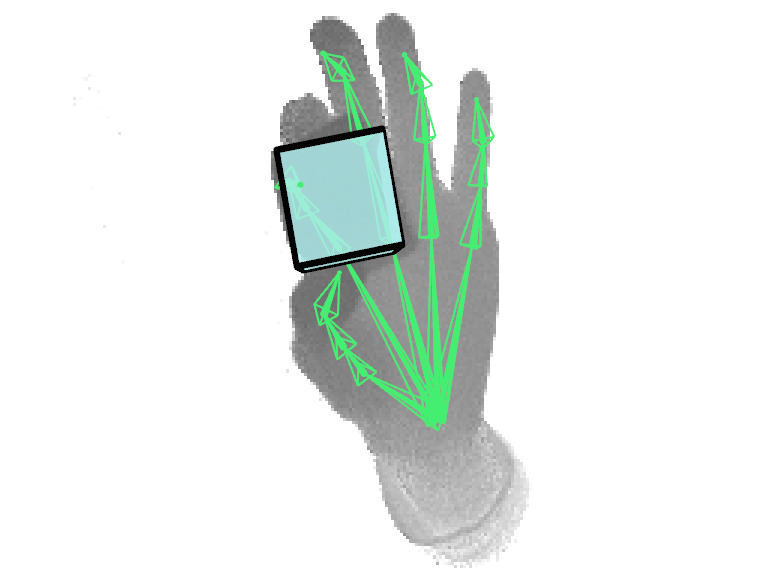}
    \includegraphics[width=0.16\linewidth]{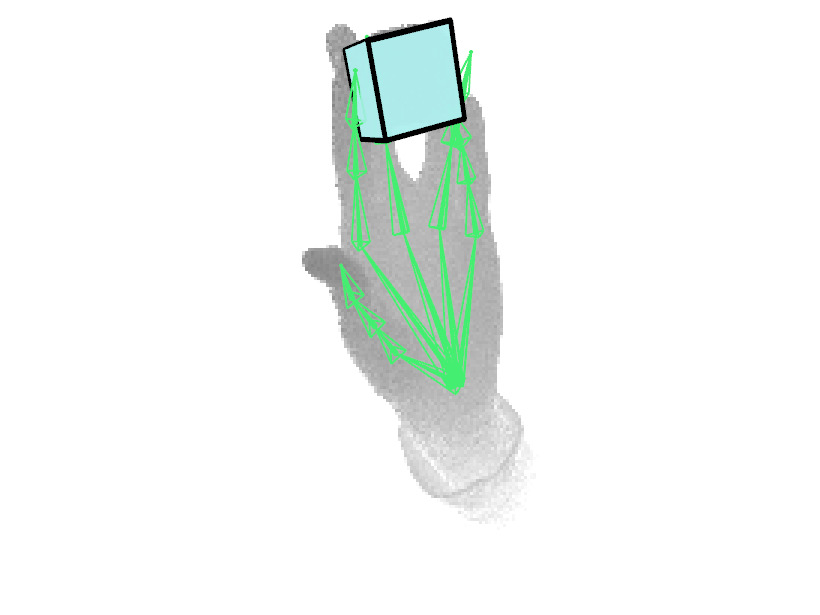}
    \includegraphics[width=0.16\linewidth]{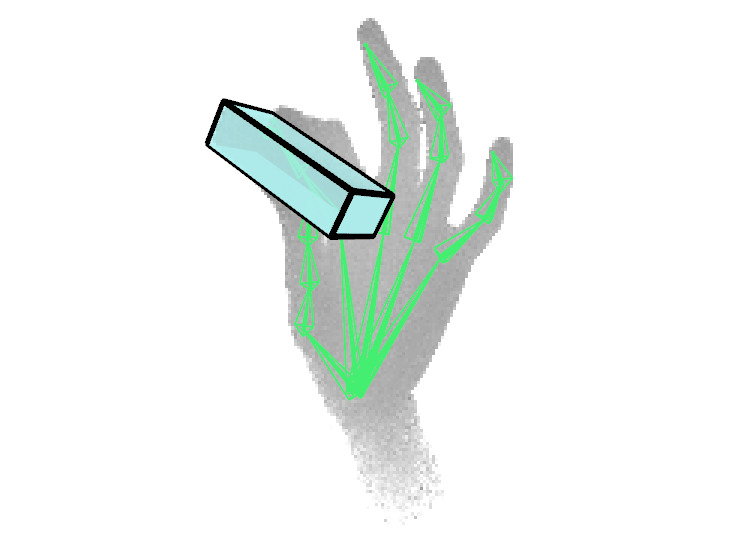}
    \includegraphics[width=0.16\linewidth]{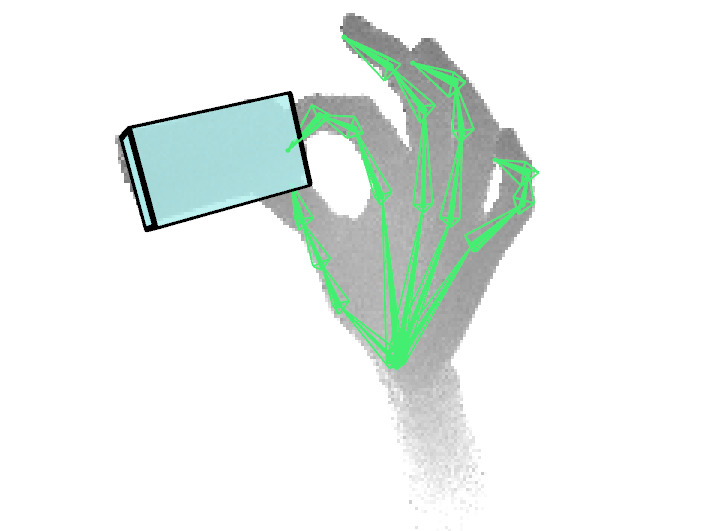}
    \includegraphics[width=0.16\linewidth]{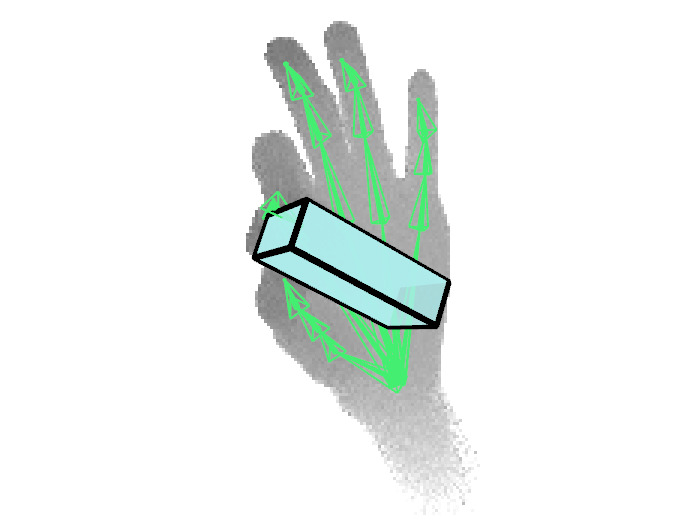}
  \caption{Real-time tracking results with various object shapes and different users}
  \end{subfigure}
  \hspace{0.2cm}	
  \begin{subfigure}[b]{\textwidth}
    \centering
    \includegraphics[width=0.19\linewidth]{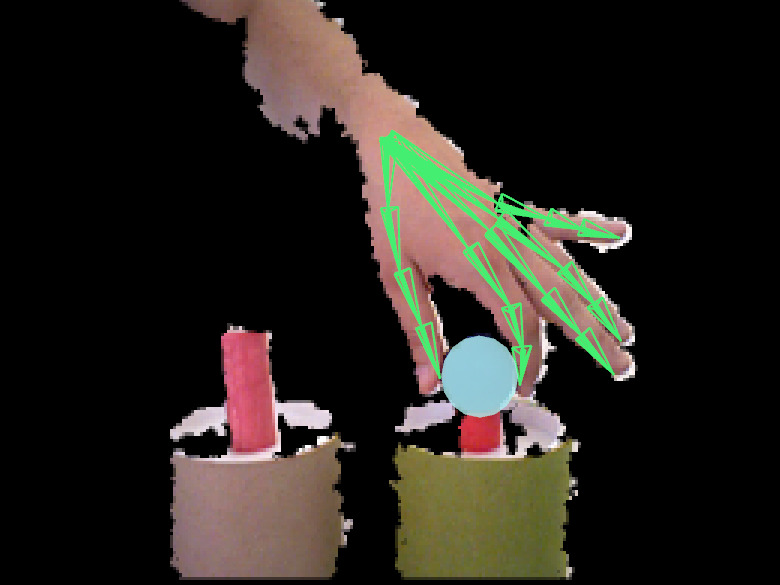}
    \includegraphics[width=0.19\linewidth]{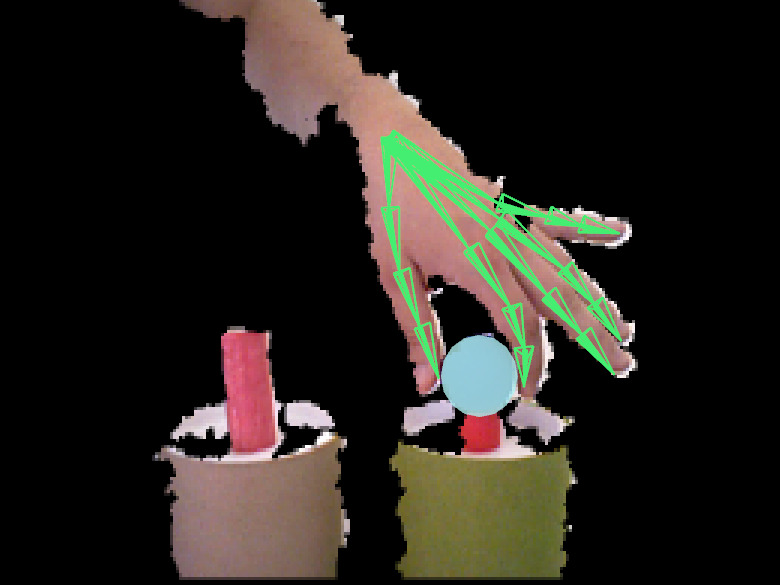}
    \includegraphics[width=0.19\linewidth]{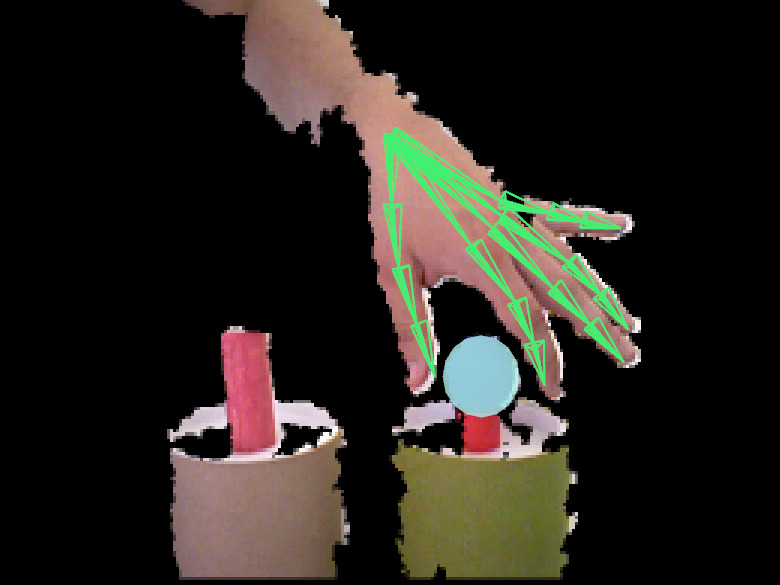}
    \includegraphics[width=0.19\linewidth]{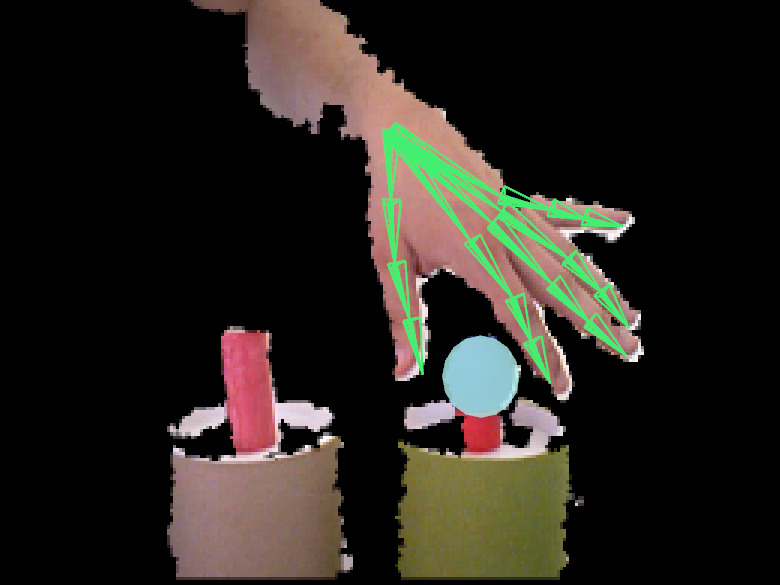}
    \includegraphics[width=0.19\linewidth]{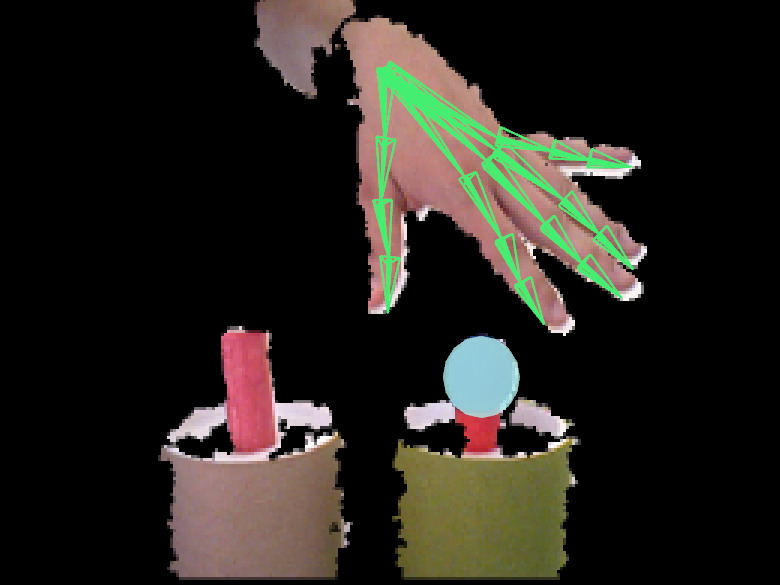}
    \includegraphics[width=0.19\linewidth]{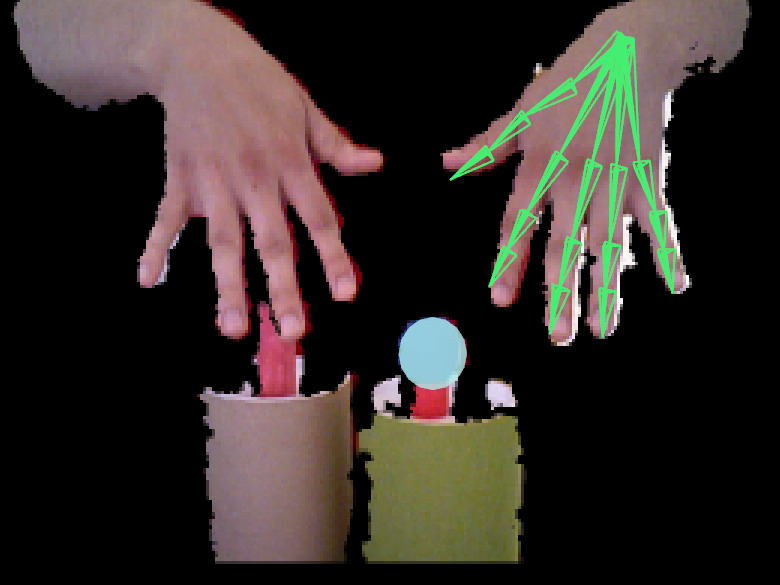}
    \includegraphics[width=0.19\linewidth]{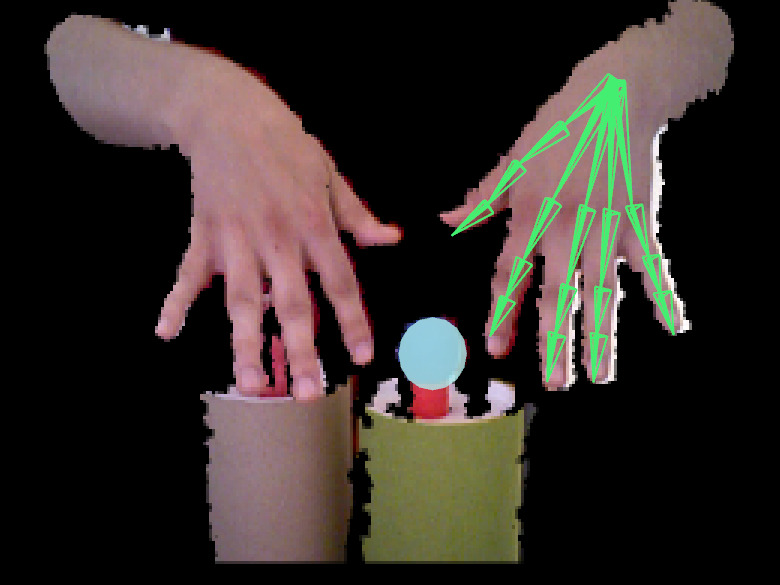}
    \includegraphics[width=0.19\linewidth]{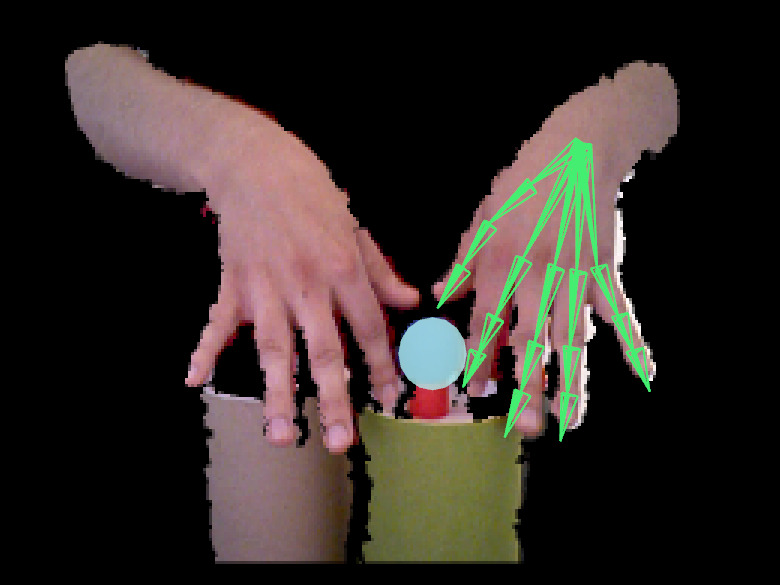}
    \includegraphics[width=0.19\linewidth]{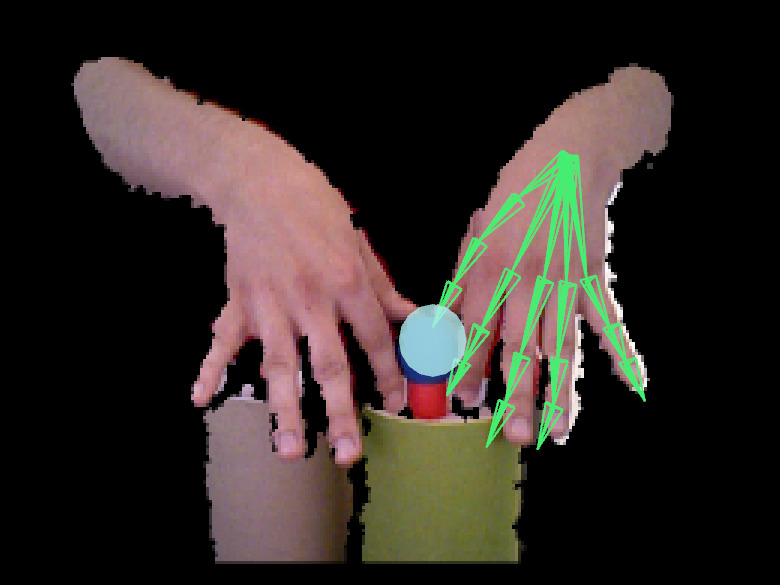}
    \includegraphics[width=0.19\linewidth]{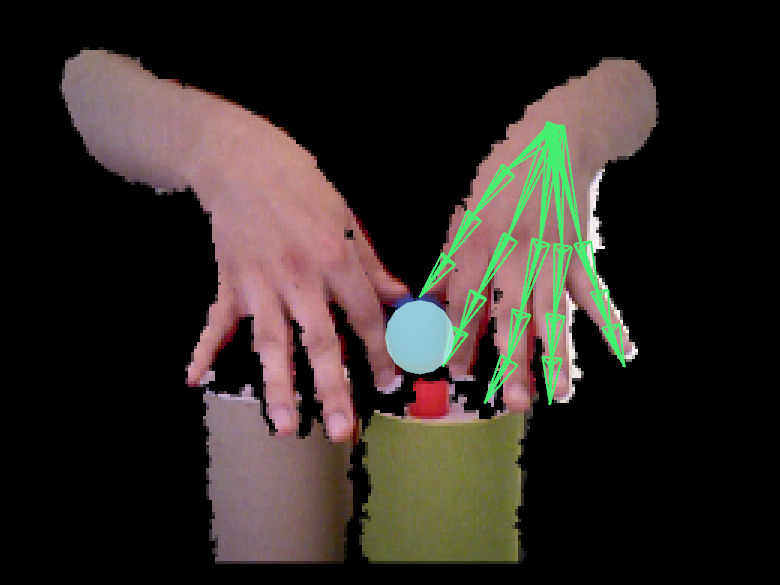}
    \includegraphics[width=0.19\linewidth]{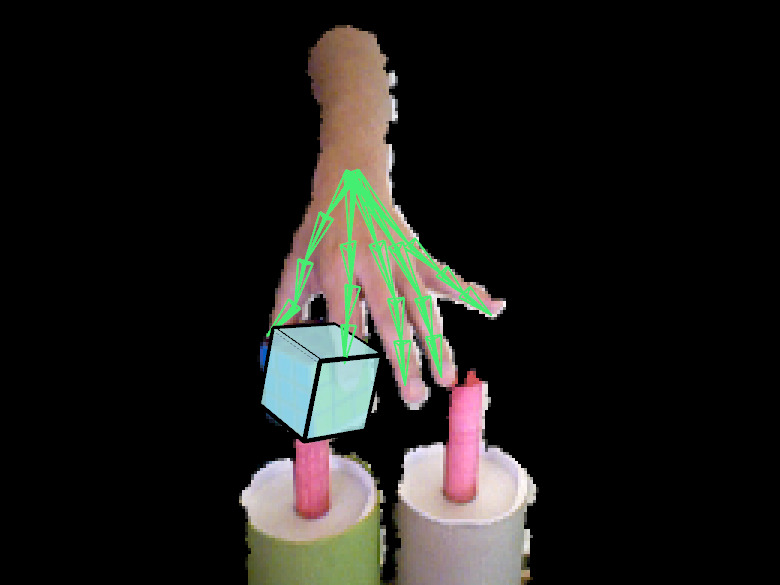}
    \includegraphics[width=0.19\linewidth]{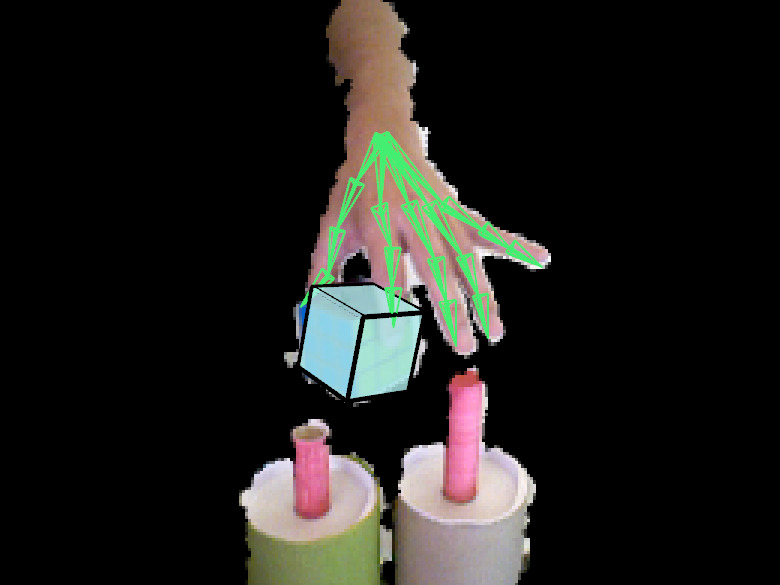}
    \includegraphics[width=0.19\linewidth]{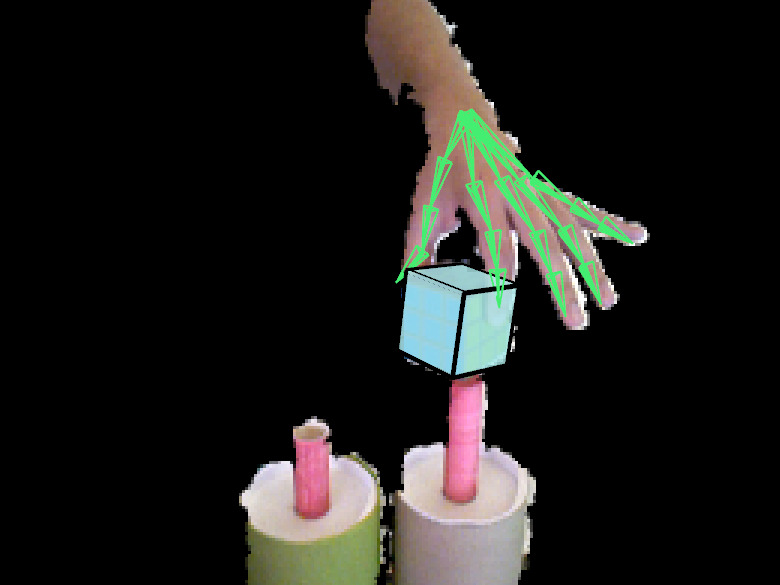}
    \includegraphics[width=0.19\linewidth]{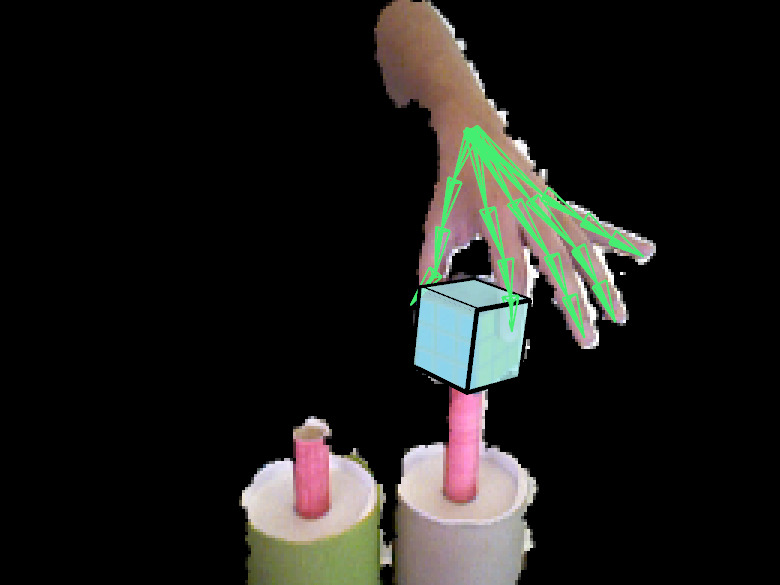}
    \includegraphics[width=0.19\linewidth]{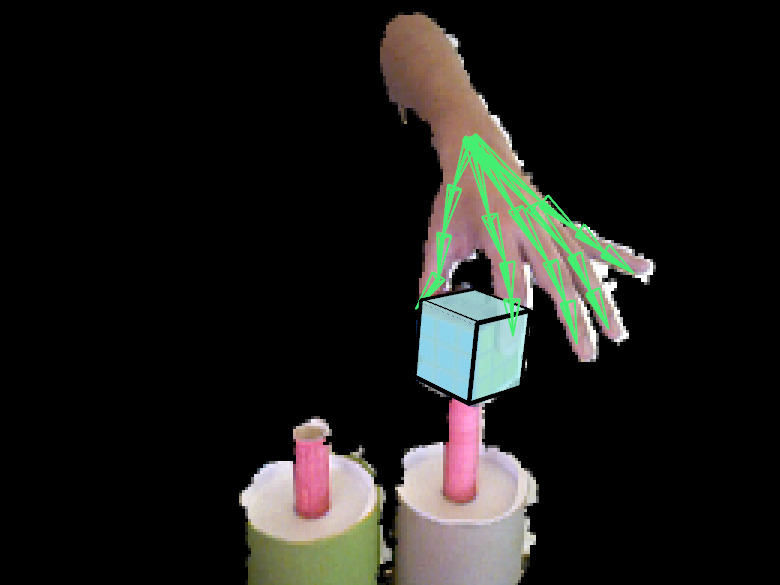}
    \caption{Results on the IJCV dataset~\cite{tzionas2015capturing}. 
	Notice how our method tracks the hand even if multiple hands are in view.
    Tracked skeleton in green and object in light blue}
  \end{subfigure}		
  \caption{(a, b) show tracking results on our dataset. (c) shows real-time results with different object shapes and colors. (d) shows results on a public dataset}
\end{figure}

\paragraph{\textbf{Limitations}} \label{sec:limitations}
%
\setlength\intextsep{0pt}
\begin{wrapfigure}{R}{0.44\textwidth}
\centering
\includegraphics[width=0.44\textwidth]{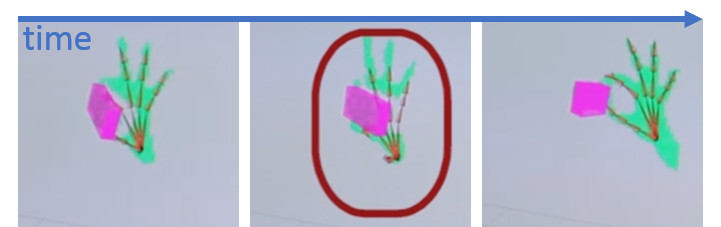}
\caption{\label{fig:occlusion_recovery}Occlusion error and recovery}
\end{wrapfigure}
%
Although we demonstrated robustness against reasonable occlusions, situations where a high fraction of the hand is occluded for a long period are still challenging.
This is mostly due to degraded classification performance under such occlusions.
Misalignments can appear if the underlying assumption of the occlusion heuristic is violated, i.\,e.\:occluded parts do not move rigidly.
Fortunately, our discriminative classification strategy enables the pose optimization to recover once previously occluded regions become visible again as shown in Fig.~\ref{fig:occlusion_recovery}.
Further research has to focus on better priors for occluded regions, for example grasp and interaction priors learned from data.
Also improvements to hand part classification using different learning approaches or the regression of dense correspondences are interesting topics for future work.
Another source of error are very fast motions. While the current implementation achieves 30~Hz, higher frame rate sensors in combination with a faster
pose optimization will lead to higher robustness due to improved temporal coherence.
We show diverse object shapes being tracked. However, increasing object complexity (shape and color) affects runtime performance.
We would like to further explore how multiple complex objects and hands can be tracked.

\section{Conclusion} \label{sec:conclusion}
We have presented the first real-time approach for simultaneous hand-object tracking based on a single commodity depth sensor.
Our approach combines the strengths of discriminative classification and generative pose optimization.
Classification is based on a multi-layer forest architecture with viewpoint selection.
We use 3D articulated Gaussian mixture alignment tailored for hand-object tracking along with novel analytic occlusion and contact handling constraints that enable successful tracking of challenging hand-object interactions based on multiple proposals.
Our qualitative and quantitative results demonstrate that our approach is both accurate and robust.
Additionally, we have captured a new benchmark dataset (with hand and object annotations) and make it publicly available.
We believe that future research will significantly benefit from this.
\\

\parahead{Acknowledgments}
This research was funded by the ERC Starting Grant projects CapReal (335545) and COMPUTED (637991), and the Academy of Finland. We would like to thank Christian Richardt.


\clearpage

\bibliographystyle{splncs03}
\bibliography{content/CVPR}
\end{document}